\pdfoutput=1

\documentclass[11pt]{article}

\usepackage{EMNLP2023}

\usepackage{times}
\usepackage{latexsym}

\usepackage[T1]{fontenc}

\usepackage[utf8]{inputenc}

\usepackage{microtype}

\usepackage{inconsolata}
\usepackage[utf8]{inputenc} 
\usepackage[T1]{fontenc}    
\usepackage{hyperref}       
\usepackage{url}            
\usepackage{booktabs}       
\usepackage{amsfonts}       
\usepackage{nicefrac}       
\usepackage{microtype}      
\usepackage{xcolor}         
\usepackage{multirow}
\usepackage{makecell}
\usepackage{graphicx}
\usepackage{epstopdf}
\usepackage{tabularx}
\usepackage{adjustbox}
\usepackage{tablefootnote}
\usepackage{listings}
\usepackage{courier}
\usepackage{amsmath}
\usepackage{listings}
\usepackage{courier}
\usepackage{amsmath}
\usepackage{arydshln}
\usepackage{graphicx}
\usepackage{dialogue}
\usepackage{blindtext}
\usepackage{subcaption}
\usepackage{enumitem}
\usepackage{float}
\usepackage{afterpage}
\definecolor{mauve}{rgb}{0.58,0,0.82}
\def\openai{\raisebox{-0.55ex}{\includegraphics[width=1.5em]{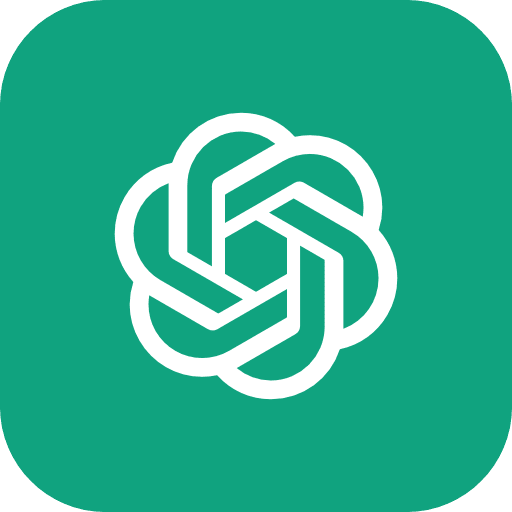}}}
\def\user{\raisebox{-0.55ex}{\includegraphics[width=1.5em]{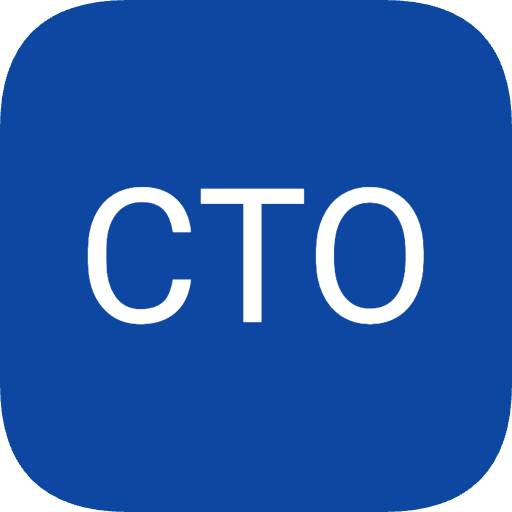}}}
\newcommand{\useblue}[1]{{\color{blue}#1}}

\title{CodeTransOcean: A Comprehensive Multilingual Benchmark\\ for Code Translation}

\author{%
  Weixiang Yan$^1$\thanks{~~Equal contribution. Work is supported by Speech Lab, Alibaba Group.}\hspace{-20mm}
  \And
  Yuchen Tian$^2$\footnotemark[1]\hspace{-20mm}
  \And
  Yunzhe Li$^3$ \hspace{-15mm}
  \And
  Qian Chen$^4$ \hspace{-5mm}
  \And
  Wen Wang$^4$ \hspace{10mm} \\
  \AND
  $^1$\textmd{University of California, Santa Barbara}
  ~~~$^2$\textmd{The University of Hong Kong} \\
  $^3$University of Illinois at Urbana-Champaign~~~$^4$\textmd{Speech Lab, Alibaba Group} \\
  \texttt{weixiangyan@ucsb.edu}~~~~~~\texttt{yuchent@connect.hku.hk} \\
  \texttt{yunzhel2@illinois.edu}~~~~~\texttt{\{tanqing.cq,w.wang\}@alibaba-inc.com} \\
}

\begin{document}
\maketitle
\begin{abstract}
Recent code translation techniques exploit neural machine translation models to translate source code from one programming language to another to satisfy production compatibility or to improve efficiency of codebase maintenance. Most existing code translation datasets only focus on a single pair of popular programming languages. To advance research on code translation and meet diverse requirements of real-world applications, we construct \textbf{CodeTransOcean}, a large-scale comprehensive benchmark that supports the largest variety of programming languages for code translation. CodeTransOcean consists of three novel multilingual datasets, namely, \textbf{MultilingualTrans} supporting translations between multiple popular programming languages, \textbf{NicheTrans} for translating between niche programming languages and popular ones, and \textbf{LLMTrans} for evaluating executability of translated code by large language models (LLMs).  CodeTransOcean also includes a novel cross-framework dataset, \textbf{DLTrans}, for translating deep learning code across different frameworks. We develop multilingual modeling approaches for code translation and demonstrate their great potential in improving the translation quality of both low-resource and high-resource language pairs and boosting the training efficiency. We also propose a novel evaluation metric \textit{Debugging Success Rate@K} for program-level code translation. Last but not least, we evaluate LLM ChatGPT on our datasets and investigate its potential for fuzzy execution predictions. We build baselines for CodeTransOcean and analyze challenges of code translation for guiding future research. 
The CodeTransOcean datasets and code are publicly available at \url{https://github.com/WeixiangYAN/CodeTransOcean}.

\end{abstract}

\section{Introduction}
\label{sec:introduction}
Early software systems are developed using programming languages such as Fortran and COBOL, which have a significantly smaller user base compared to modern mainstream programming languages (e.g., Python and Java). Hence maintaining and modernizing early software systems are expensive~\cite{freecodecamp2020}. Moreover, the readability and compatibility of the mixed multitude of programming languages are challenging when migrating existing software systems to new technology ecosystems or integrating software systems using different programming languages. The code translation task aims to convert source code from one programming language to another and is of great value in industry.  

\begin{table*}
    \begin{adjustbox}{width=\textwidth}
    \renewcommand{\arraystretch}{1.0}
    \begin{tabular}{cccccc}
    \toprule
        \textbf{Category} & \textbf{Language/Framework} & \textbf{Dataset Name} & \textbf{Train/Dev/Test \#Samples} & \textbf{Avg. \#Tokens/Sample} & \textbf{Avg. Length}\\
    \midrule
    \multirow{8}{*}{Multilingual} & \makecell*[c]{Python, C, C++, \\Visual Basic, Go,\\ PHP, Java, C\# } & MultilingualTrans & 19,115~/~3,759~/~7,545 & 398~/~421~/~491 & 1099~/~1135~/~1358 \\
    \cline{2-6}
    & \makecell*[c]{Swift, R, Rust, \\Fortran, Ada, Perl,\\ COBOL, Lua, ... } & NicheTrans & 165,457~/~23,509~/~47,502~ & 292~/~375~/~505 & 785~/~995~/~1372 \\
    \cline{2-6}
    & \makecell*[c]{Python, C, C++, \\Visual Basic, Go,\\ PHP, Java, C\# } & LLMTrans & --~/~--~/~350~ & --~/~--~/~270~ & --~/~--~/~745~ \\
    \midrule
    Cross-Framework & \makecell[c]{PyTorch, TensorFlow,\\ MXNet, Paddle} & DLTrans  & 282~/~36~/~90  & 625~/~1102~/~875 & 1318~/~2441~/~1841 \\  
    \bottomrule
  \end{tabular}
  \end{adjustbox}
    \caption{Summary of our \textbf{CodeTransOcean}. We report \#Samples, Avg. \#Tokens/Sample and Avg. Length for Train/Dev/Test sets of each dataset. Note that LLMTrans is only for testing. \#Samples are on the \textbf{program-level}. \#Tokens are based on RoBERTa tokenizer~\cite{DBLP:journals/corr/abs-1907-11692}. Length is the number of characters.}
      \label{table:CodeTransOcean}
\end{table*}

Code translation methods evolve from the inefficient, costly, and error-prone manual rewriting method to automatic methods. Automatic code translation methods can be categorized into \textit{compilers and transpilers}, \textit{rule-based methods}, and \textit{neural network based methods}. Neural models~\cite{feng2020codebert,wang2021codet5,wang2023codet5} have become dominant in code translation. Details of code translation methods are presented in Appendix~\ref{appendix:related_work}. The performance of neural models relies heavily on large-scale high-quality parallel data. However, existing code translation datasets are limited by \textbf{insufficient coverage of programming languages and mostly focusing on a single pair of popular programming languages}, \textbf{limited scale}, and \textbf{uneven data distribution}. The widely used CodeTrans~\cite{lu2021codexglue} is a small dataset containing only Java-C\# parallel data for quite short code samples. Other datasets~\cite{ahmad2021avatar, lachaux2020unsupervised, zhu2022multilingual, nguyen2013lexical, chen2018tree} suffer from the same limitations. Consequently, existing code translation models~\cite{feng2020codebert,wang2021codet5,ahmad2021unified} are confined to a narrow range of one-to-one code translation scenarios.
Moreover, deep learning has been broadly used and achieved unprecedented success. However, there are barriers between different deep learning frameworks during the actual production process. Existing code translation datasets also neglect important demands from real-world applications, including \textbf{modernizing early software systems developed in \textit{niche} programming languages} and \textbf{migrating code across different deep learning frameworks}. 

To address these limitations and advance neural code translation models, we construct a large-scale comprehensive multilingual code translation benchmark \textbf{CodeTransOcean}, summarized in Table~\ref{table:CodeTransOcean}. CodeTransOcean is an innovative benchmark that aims to provide a \textbf{unified platform} for evaluating various models \textit{on a comprehensive set of code translation tasks} that reflect real-world demands.  Based on this goal, each dataset in CodeTransOcean is specifically designed to tackle a key challenge in the field of code translation. CodeTransOcean includes three \textit{multilingual} datasets, namely, the \textbf{MultilingualTrans} dataset (including eight popular programming languages), the \textbf{NicheTrans} dataset (translating between thirty-seven niche programming languages and the eight popular ones\footnote{We define popular and niche programming languages based on the TIOBE Programming Community Index, which is a metric of the popularity of programming languages.}), and a specialized dataset \textbf{LLMTrans} (including 350 data samples and their executed results) to evaluate executability of code translated by large language models (LLMs), and a \textit{cross-framework} dataset \textbf{DLTrans} facilitating our proposed task for translating code between deep learning frameworks to enhance code reusability. DLTrans includes 408 samples covering four mainstream deep learning frameworks. 

Multilingual modeling shows great potential in neural machine translation~\cite{aharoni2019massively,wang2020multitask,zhu2023multilingual}, but it has not been systematically explored for code translation. We investigate multilingual modeling for code translation using our MultilingualTrans, NicheTrans, and DLTrans datasets. Experimental results demonstrate that multilingual modeling significantly improves translation quality for both \textit{high-resource} and \textit{low-resource} language pairs and improves the model training efficiency.

Recent research indicates that the proficiency of the LLM ChatGPT in natural language translation is on par with commercial-grade translation systems~\cite{jiao2023chatgpt}. \textbf{To the best of our knowledge, our work is the first to systematically investigate the potential of ChatGPT in code translation}. We develop a fully automated translation-execution-evaluation pipeline \textbf{AutoTransExecuter} to support this study. Note that \textit{match-based metrics} and \textit{execution-based metrics} have been used for evaluating code translation methods, with details in Appendix~\ref{appendix:related_work}. In order to accurately evaluate the usability of translated code from ChatGPT, we propose a novel execution-based evaluation metric \textbf{Debugging Success Rate @K (DSR@K)}, which is the percentage of samples with translation results that successfully execute and produce the expected functionality after $K$ debugging rounds. On our LLMTrans dataset, the baseline ChatGPT setting achieves 48.57\% DSR@0. We find that self-debugging and one-shot improve the performance while chain-of-thought strategies degrade the translation accuracy. Since our AutoTransExecuter still cannot cover arbitrary programming languages, we also propose a novel metric \textit{fuzzy execution}, attempting to address the limitations of existing evaluation metrics for code translation. Our preliminary study using ChatGPT shows that ChatGPT is still inadequate to predict fuzzy execution for any arbitrary programming language, which demands future research.

Our contributions can be summarized as follows:
\begin{itemize}[leftmargin=*,noitemsep]
\item \textbf{A large-scale multilingual code translation  benchmark}: CodeTransOcean covers the largest number of popular and niche programming languages so far with the largest scale. It also includes an unprecedented dataset for translating code across different deep learning frameworks and a dataset and an automated pipeline for evaluating LLMs on code translation. We establish baselines for all datasets in CodeTransOcean.
 
\item \textbf{Multilingual modeling for code translation}: We are the first to systematically evaluate multilingual modeling on code translation for both high-resource and low-resource language pairs. Experimental results demonstrate that multilingual modeling significantly improves translation quality for both \textit{high-resource} and \textit{low-resource} language pairs and improves training efficiency.
 
\item \textbf{ChatGPT on code translation}: We conduct the first comprehensive study of the potential of ChatGPT on code translation, investigating efficacy of prompting strategies, hyperparameters, self-debugging, One-shot, and Chain-of-Thought.
 
\item \textbf{New evaluation metrics}: We propose \textit{DSR@K} to evaluate translation and debugging capabilities of LLMs. We also propose a \textit{fuzzy execution} metric based on LLMs and conduct a preliminary study using ChatGPT on this metric.
\end{itemize}

\section{Related Work}
\label{related_work}

\paragraph{Code Translation Datasets}~~The success of neural models for code translation relies heavily on large-scale high-quality parallel data. However, existing code translation datasets are plagued by issues such as \textit{insufficient coverage of programming languages}, \textit{limited scale}, and \textit{imbalanced data distribution}. The widely used code translation dataset CodeTrans~\cite{lu2021codexglue} in the CodeXGLUE benchmark consists of Java-C\# function pairs. The small parallel corpus AVATAR~\cite{ahmad2021avatar} is constructed for Java-Python code translation. \citet{nguyen2013lexical} construct a Java-C\# dataset to explore statistical machine translation on code translation tasks\footnote{It was not possible to count specific information about this dataset because it was not released to the public and we were unable to obtain response from the authors.}. \citet{chen2018tree} explore this dataset from Nguyen et al. and also construct a CoffeeScript-JavaScript parallel dataset for investigating tree-to-tree neural models for code translation. \citet{lachaux2020unsupervised} create a dataset containing 852 programs to evaluate unsupervised methods. Recently, \citet{zhu2022multilingual} construct a new translation dataset CoST from the GeeksForGeeks website\footnote{In Table \ref{table:CTDatasets}, we report the \textbf{program-level} counts for the CoST dataset to facilitate a fair comparison with our own program-level datasets.}. Subsequently, they release the translation dataset XLCoST~\cite{zhu2022xlcost}, which expands the CoST dataset by 7.3 times. However, the limited language coverage of these datasets and their imbalanced data distribution hinder their practical applications. \citet{roziere2022leveraging} construct the TransCoder-ST dataset to perform unsupervised code translation using automated unit tests. Details of these datasets are summarized in Table~\ref{table:CTDatasets}. \citet{10055851} proposes a code translation dataset XTest containing nine programming languages with unit tests, but it is not open-sourced\footnote{We tried to contact the authors but there was no response.}. Although CodeNet \cite{puri2021project} comprises many problem statements and provides corresponding solutions, experts have proven that about half of the CodeNet dataset has incorrect solutions~\cite{zhu2022multilingual}, making it unsuitable for code translation tasks. With the limitations of existing code translation datasets, neural models trained on them may encounter overfitting, underfitting, and poor generalizability. Clearly, these issues impede the development of neural models for code translation. Therefore, constructing datasets that effectively address these problems is critical to enhance performance of code translation algorithms. 

\paragraph{Code Translation Methods and Evaluation Metrics}~~Details of code translation methods and evaluation metrics are presented in Appendix~\ref{appendix:related_work}.

\begin{table*}
  \begin{adjustbox}{width=\textwidth}
    \renewcommand{\arraystretch}{1.0}
  \begin{tabular}{ccccc}
    \toprule
    \textbf{Dataset Source} & \textbf{Programming Languages} & \textbf{\#Samples} & \textbf{Avg. \#Tokens/Sample} & \textbf{Avg. Length} \\
    \midrule
    CodeTrans~\cite{lu2021codexglue}  & Java, C\# & 11,800 & 59~/~63~/~58 & 205~/~218~/~202 \\  
    Avatar~\cite{ahmad2021avatar} & Java, Python & 9,517 &  239~/~235~/~234 & 691~/~687~/~688  \\  
    Nguyen et al.\cite{nguyen2013lexical} & Java, C\# & 16,966  & -- & -- \\  
    Lachaux et al.\cite{lachaux2020unsupervised} & C++, Java, Python & 852  &  -~/~119~/~120 & -~/~313~/~311  \\  
    CoST~\cite{zhu2022multilingual} & \makecell[c]{C++, Java, Python, C\#,\\ Javascript, PHP, C} & 16,738 &  272~/~180~/~199 & 770~/~458~/~511   \\ 
    TransCoder-ST~\cite{roziere2022leveraging} & Java, C++, Python & 437,030 &  -- & -- \\
    XLCoST~\cite{zhu2022xlcost} & \makecell[c]{C++, Java, Python, C\#,\\ Javascript, PHP, C} & 122,151  &  234~/~232~/~222 & 644~/~634~/~606   \\ 

    \bottomrule
  \end{tabular}
    \end{adjustbox}
  \caption{Summary of existing code translation datasets. For \#Samples, we report \textbf{program-level} counts for Avatar, CoST, and XLCoST. Given that the original samples from other datasets are not organized at the program-level, we report counts at the snippet-level for these datasets. Avg. \#Tokens/Sample and Avg. Length are counted in the same way as Table~\ref{table:CodeTransOcean}.}
    \label{table:CTDatasets}
\end{table*}

\section{The CodeTransOcean Benchmark}
\label{CTOBenchmark}

In this section, we provide detailed descriptions and analyses of our CodeTransOcean benchmark, including the code translation tasks, their associated datasets, and dataset statistics. Details of data collection methods and licensing information as well as quality control and quality assessment are presented in Appendix~\ref{sec:data_management}. 
Note that the vast majority of the samples in CodeTransOcean provides explicit input and output, which is equivalent to unit tests. 
Overall, CodeTransOcean consists of 270,507 samples (over 200K unit tests), covering 45 programming languages for multilingual code translation and 4 deep learning frameworks for cross-framework code translation\footnote{Code Translation also extends to conversions between different versions of the same language, e.g., Python 2 to Python 3. However, according to our survey, these translation tasks are quite straightforward. Naive Copy methods, specific translation tools, and tutorials (e.g., \href{https://docs.python.org/2/library/2to3.html}{Python 2 to 3 Conversion Guide}) already achieve high translation accuracy. As a result, we no longer include these types of tasks in our benchmark.}. Note that all samples in all CodeTransOcean datasets are constructed at the \textbf{program-level}. We ensure a balanced distribution of each language/framework when constructing the datasets (Appendix~\ref{sec:data_management}). There is no overlap between CodeTransOcean datasets and existing code translation datasets.

\subsection{Multilingual Code Translation}
\label{subsec:multilingual}
With the increasing need to unify the language variety when implementing system integration or extensions with multilingual programming environments, we construct the MultilingualTrans dataset for multiple popular programming languages\footnote{We categorize languages as popular or niche based on the TIOBE Index Programming Language Rankings released in April 2023 \url{https://www.tiobe.com/tiobe-index/}.}. Among programming languages in the rankings, we select the Top-10 languages as popular ones except JavaScript and SQL\footnote{It is important to note that JavaScript and SQL, both within the top 10, are mainly used for front-end programming and database management respectively, signifying considerable differences in their usage scenarios compared to the other 8 languages.} and construct the MultilingualTrans dataset based on the 8 programming languages. We treat the other languages in the rankings as niche languages and construct the NicheTrans dataset for translating between niche languages and popular languages.  Additionally, in order to quantitatively evaluate the execution capabilities of the code generated by LLMs (e.g., ChatGPT, PaLM2~\cite{anil2023palm}), we construct LLMTrans, which includes the execution results for a subset of MultilingualTrans and facilitates evaluating LLMs for multilingual code translation. 

\paragraph{MultilingualTrans Dataset} 
\label{sec:multilingualTrans_dataset} 
This dataset contains 30,419 program samples covering eight popular programming languages, namely,  C, C++, C\#, Java, Python, Go, PHP, and Visual Basic. Table~\ref{table:Multilingual} shows the statistics of each language pair. Note that XLCoST \cite{zhu2022xlcost} is the only existing multilingual code translation dataset. Compared to XLCoST, MultilingualTrans is advantageous in more balanced data distribution across various programming languages, practicality of language pairs, and data quality. For example, the real-world requirement for translating Java into JavaScript as in XLCoST is quite limited. As to data quality, our MultilingualTrans originates from a programming chrestomathy website, with all data already reviewed and verified by the website.

\paragraph{NicheTrans Dataset} 
\label{sec:nicheTrans_dataset} 
The NicheTrans dataset contains 236,468 program samples, covering code translation pairs from thirty-seven niche programming languages, including Ada, COBOL, Pascal, Perl, Erlang, Fortran, Scala, Julia and others, to the eight popular ones. Table~\ref{table:NicheTrans} shows statistics of each niche language. Although many studies have highlighted the practical necessity of code translation for modernizing niche programming languages \cite{chen2018tree, zhu2022multilingual,lachaux2020unsupervised}, our NicheTrans dataset is the first dataset for code translation between these niche languages and popular ones. We believe this dataset will not only facilitate modernization of outdated programming languages more effectively, but also augment and evaluate generalizability of neural models.

\paragraph{LLMTrans Dataset} The LLMTrans dataset aims to provide a benchmark for evaluating the performance of LLMs on code translation. The dataset translates seven popular programming languages to Python, totaling 350 program samples. We compile and test these samples and record the execution results. Based on this dataset, we design and implement an automated pipeline, \textbf{AutoTransExecuter}\footnote{AutoTransExecuter only supports translation from any source language to Python. We discuss it in \hyperref[sec:limitations]{Limitations}.}, automatically using LLMs to conduct code translation, execution, debugging, and calculating the success rate. This dataset and the automated pipeline ease investigation of the actual debugging success rate of LLMs on code translation and effectively measure the practical usability of LLMs.  Details of the LLMTrans dataset are in Table~\ref{table:CodeTransOcean}.

\afterpage{
\begin{table*}[!ht]
\centering
\begin{adjustbox}{width=0.7\textwidth}
\begin{tabular}{ccccc}
\toprule
 \textbf{Average} & \textbf{One-to-One (baseline)} & \textbf{Many-to-One} & \textbf{Many-to-Many} & \textbf{One-to-Many} \\ 
\midrule
High-resource & 4.68 & 5.56 ($\uparrow$ 0.88) & 5.94 ($\uparrow$ 1.26) & 6.18 ($\uparrow$ 1.50) \\
Low-resource  & 4.83 & 4.85 ($\uparrow$ 0.02) & 4.95 ($\uparrow$ 0.12) & 5.84 ($\uparrow$ 1.01) \\
All & 5.19 & 5.31 ($\uparrow$ 0.12) & 5.81 ($\uparrow$ 0.62) & 6.42 ($\uparrow$ 1.23) \\
\bottomrule
\end{tabular}
\end{adjustbox}
\caption{Average BLEU scores of the four multilingual modeling strategies, \textbf{One-to-One}, \textbf{Many-to-One}, \textbf{Many-to-Many}, and \textbf{One-to-Many}, for All language pairs, High-resource language pairs, and Low-resource language pairs.}
\label{tab:high_low_resource}
\end{table*}
}

\subsection{Cross-framework Code Translation}
\label{sec:DLTask}

\paragraph{Cross-Deep-Learning-Framework Translation Task} 
The widespread applications of deep learning (DL) has spawned emergence of various DL frameworks, such as PyTorch, TensorFlow, MXNet, and Paddle. However, there are significant differences in syntax and dependency libraries between different frameworks, severely impeding reusability of projects\footnote{\url{https://www.assemblyai.com/blog/pytorch-vs-tensorflow-in-2023/}}. Moreover, studies illustrate significant disparities in energy consumption and economic costs during training and inference between various frameworks~\cite{georgiou2022green}. Selecting an appropriate DL framework for green AI has become paramount in an era of large models~\cite{anil2023bigger}. 
Code reusability and energy-economic efficiency in DL have emerged as critical considerations for both research and practical engineering implementation. Converting code between different DL frameworks is challenging, mainly due to differences between frameworks, code complexity, structural inconsistencies, and cross-platform compatibility (more details are in Appendix~\ref{appendix:specific_challenges}). Existing cross-DL-framework adaptive technologies such as the ONNX\footnote{\url{https://onnx.ai/}} model conversion protocol require both parties to import and export based on agreed data formats or to convert only the final model through the computation graphs. These technologies have obvious limitations. 
In contrast, we propose a \textbf{Cross-DL-framework Translation} task for code migration between different DL frameworks through code translation (Appendix~\ref{appendix:dl_examples}). Compared to existing cross-framework adaptive technologies, Cross-DL-framework Translation achieves re-implementation under multiple DL frameworks through an automated process, which not only generates highly readable code and enables secondary development, but also provides developers with flexibility on combining advantages of multiple frameworks. 

\paragraph{DLTrans Dataset}
We construct the \textbf{DLTrans dataset} for Cross-DL-framework Translation, including four deep learning frameworks and spanning twelve directions. To the best of our knowledge, our work is the first to define the cross-DL-framework translation task and construct a corresponding dataset. We create two subsets of different granularities based on the collected code, namely, \textit{coarse-grained} at the program level and \textit{fine-grained} at the function or class level. Each code pair comprises code that shares the same functionality but is written in different popular DL frameworks, including PyTorch, TensorFlow, MXNet, and Paddle. The coarse-grained and fine-grained datasets have 
408 and 3,270 samples, respectively. In this work, we only experiment on the coarse-grained subset.

\section{Experiments}
We present experiments of multilingual training for code translation (Section~\ref{sec:multilingual_modeling}). We then introduce a novel evaluation metric \textbf{Debugging Success Rate@K} for \textbf{program-level} code translation (Section~\ref{sec:csr}) and the first comprehensive exploration of ChatGPT for code translation (Section~\ref{sec:chatgpt}).

\subsection{Multilingual Modeling}
\label{sec:multilingual_modeling}

Multilingual modeling has been pivotal in broadening the applicability of neural machine translation \cite{aharoni2019massively,wang2020multitask,zhu2023multilingual,johnson2017googles}. This is primarily evidenced in enhancing the performance of low-resource languages and cross-language transfer learning~\cite{mohammadshahi2022small100,zoph2016transfer,nguyen2017transfer,johnson2017googles}. CodeTransOcean covers nearly fifty programming languages and deep learning frameworks. We use its datasets to explore multilingual modeling on code translation tasks.

\paragraph{Experimental Setups}
In this work, we use pre-trained CodeT5+~\cite{wang2023codet5}\footnote{We will conduct evaluations of a broader selection of models on our datasets in future work, including LLaMA~\cite{touvron2023llama}, WizardCoder~\cite{luo2023wizardcoder}, etc.} as the backbone based on its superior performance on code understanding and generation evaluations reported in~\cite{wang2023codet5}. We use the MultilingualTrans dataset to investigate four multilingual modeling strategies based on data sharing in the source or target language or both, namely,  \textit{One-to-One}, \textit{One-to-Many}, \textit{Many-to-One}, and \textit{Many-to-Many}, with One-to-One as the baseline. Details of the four strategies are in Appendix~\ref{appendix:multilingual_modeling}.
To understand the strengths and weaknesses of the four strategies, we compare their average performance on \textit{all language pairs} and focus on \textit{low-resource} and \textit{high-resource pairs}. Since the CodeBLEU metric~\cite{ren2020codebleu} does not cover all eight languages in MultilingualTrans, we use BLEU to measure translation accuracy for the four strategies. Then, we establish baselines for the DLTrans and NicheTrans datasets.

We rank the resource richness of the eight programming languages in MultilingualTrans in descending order based on their amounts in the CodeT5+ pre-training data, as Java, PHP, C, C\#, Python, C++, and Go (Visual Basic is not covered by the CodeT5+ pre-training data). Based on this ranking, we consider Visual Basic, C++, and Go as low-resource languages and Java, PHP and C as high-resource languages.

\afterpage{
\begin{table*}
\centering
  \begin{adjustbox}{width=\textwidth}
\begin{tabular}{cccccccccccccc}
\toprule
\multirow{2}{*}{\textbf{Method}} & \multirow{2}{*}{\textbf{Metric}}   & \multicolumn{3}{c}{\textbf{PyTorch}} & \multicolumn{3}{c}{\textbf{TensorFlow}} & \multicolumn{3}{c}{\textbf{MXNet}} & \multicolumn{3}{c}{\textbf{Paddle}} \\ 
\cmidrule(lr){3-5} \cmidrule(lr){6-8} \cmidrule(lr){9-11} \cmidrule(lr){12-14}
  &  & EM & BLEU & CodeBLEU & EM & BLEU & CodeBLEU & EM & BLEU & CodeBLEU & EM & BLEU & CodeBLEU \\
     \midrule
\multirow{4}{*}{Naive} & PyTorch   & -- & -- & -- & 27.27 & 66.25  & 69.46  & 28.18 & 72.77 & 76.63 & 30.91 & 80.35 & 83.13 \\ 
& TensorFlow   & 27.27 & 66.32 & 68.92 & -- & -- & -- & 29.09 & 63.79 & 67.94 & 27.27 & 63.04 & 65.81 \\ 
& MXNet  & 28.18 & 72.86 & 74.15 & 29.09 & 63.84 & 66.06  & -- & -- & -- & 28.18 & 69.49 & 71.09 \\   
& Paddle  & 30.91 & 80.25 & 84.83 & 27.27 & 62.94  & 67.78  & 28.18 & 69.43 & 75.09 & -- & -- & -- \\ 
    \midrule
\multirow{4}{*}{CodeT5+} & PyTorch   & -- & -- & -- & 35.45$\pm$0.91 & 71.16$\pm$0.73  & 70.54$\pm$0.75  & 42.73$\pm$2.41 & 81.76$\pm$0.45 & 82.52$\pm$0.56 & 43.64$\pm$1.58 & 85.76$\pm$0.60 & 85.07$\pm$0.74 \\ 
& TensorFlow   & 34.85$\pm$1.38 & 71.97$\pm$0.56 & 71.08$\pm$0.72 & -- & --  & --  & 36.67$\pm$1.89 & 72.77$\pm$0.61 & 73.04$\pm$0.18 & 29.70$\pm$2.63 & 69.38$\pm$0.38 & 68.76$\pm$0.32 \\  
& MXNet   & 32.12$\pm$2.29 & 77.79$\pm$0.13 & 76.43$\pm$0.14 & 31.82$\pm$1.58 & 67.22$\pm$0.39  & 67.68$\pm$0.27  & -- & -- & -- & 29.09$\pm$0.91 & 74.26$\pm$0.46 & 73.27$\pm$0.42 \\  
& Paddle   & 43.03$\pm$4.10 & 86.25$\pm$0.86 & 86.09$\pm$0.88 & 29.39$\pm$2.93 & 69.43$\pm$0.57  & 69.57$\pm$0.51  & 35.75$\pm$0.53 & 78.65$\pm$0.62 & 79.46$\pm$0.38 & -- & -- & -- \\  
\bottomrule
\end{tabular}
  \end{adjustbox}
\caption{Results on DLTrans of Naive and CodeT5+\_220M with \textbf{Many-to-Many} strategy. We run each experiment with 3 random seeds and report the mean and standard deviation of EM, BLEU, and CodeBLEU scores.}
\label{tab:multilingual_modeling_dl}
\end{table*}
}

\afterpage{
\begin{table}
\centering
\begin{adjustbox}{width=0.4\textwidth}
\begin{tabular}{ccccccc}
\toprule
 \textbf{BLEU} & \textbf{Naive} & \textbf{Two-way} & \textbf{One-way} \\ 
     \midrule
Many-to-C & 2.36 & 4.60 & 4.86 \\
Many-to-C\# & 2.53 & 4.48 & 3.82  \\
Many-to-C++ & 1.99 & 4.78 & 3.32  \\
Many-to-Go & 3.11 & 5.24 & 3.19 \\
Many-to-Java & 3.18 & 5.23 & 5.34  \\
Many-to-PHP & 4.37 & 2.46 & 1.98 \\
Many-to-Python & 2.87 & 2.38 & 1.67  \\
Many-to-VB & 1.69 & 2.17 & 1.97  \\
\hdashline
Average & 2.76 & \textbf{3.92} & 3.27 \\
\bottomrule
\end{tabular}
 \end{adjustbox}
 \caption{BLEU scores on NicheTrans of Naive and CodeT5+\_220M with \textbf{Many-to-Many} strategy. \textbf{One-way} denotes training models only from niche to popular, while \textbf{Two-way} denotes training in both directions.}
\label{tab:multilingual_modeling_niche}
\end{table}
}

\paragraph{Results and Analysis} \label{sec:multilingual_modeling_analysis}
Detailed experimental results are shown in Table~\ref{tab:multilingual_modeling_strategies} in Appendix. For \textbf{All} language pairs, the performance of the four strategies is ranked as \textbf{One-to-Many > Many-to-Many > Many-to-One > One-to-One}. (1) Under One-to-Many strategy, the model encoder can provide more comprehensive information for source language translation due to its ability to absorb more source language features, thereby improving generalizability of the model. (2) Many-to-Many can be considered as expanding the One-to-Many strategy by employing a greater volume of non-source language data for training. Since the encoder must be attuned to the features of various languages simultaneously under Many-to-Many, parameter sharing may potentially undermine the performance.
(3) Many-to-One helps the model to learn from a broader range of data than the baseline. Specific patterns or expressions in diverse source languages assist the model in more precisely comprehending how to translate into the target language. The shared semantic representations across different source languages allow the model to implement effective transfer learning strategies. Furthermore, increase in training samples enables the model to optimize the loss function more stably. These results are consistent with previous findings on multilingual modeling for natural language translation~\cite{aharoni2019massively}: Many-to-Many models, trained across multiple target languages instead of just one target language, can function effectively as a regularization strategy for Many-to-One, thereby reducing the possibility of over-matching.

For \textit{High-resource} and \textit{Low-resource} languages, as shown in Table~\ref{tab:high_low_resource}, the ranking of the four strategies is the same as for \textit{All}, but there is notable difference in their adaptability across languages of varying resource scales. High-resource languages can take advantage more effectively from the shared information across multiple source languages; whereas, low-resource languages are relatively less equipped to handle the additional uncertainty and noise introduced by shared parameters, and thus often have to rely on a larger volume of source language data to optimize their benefits.

Results from the Many-to-Many strategy on DLTrans and NicheTrans datasets are shown in Tables~\ref{tab:multilingual_modeling_dl} and \ref{tab:multilingual_modeling_niche}. The experimental results suggest that significant improvements in translation accuracy can be achieved by swapping the source and target languages in the training set to facilitate data augmentation and training a bidirectional model.

Notably, prior studies on multilingual neural machine translation often overlook the comparison between One-to-Many and other strategies. Nevertheless, One-to-Many demonstrates superiority over the One-to-One baseline across all our experiments. Overall, our results strongly recommend a targeted multilingual modeling strategy for code translation, as it not only can translate multiple language pairs with a single model, but also achieves better and more stable accuracy than baselines.

\subsection{Debugging Success Rate@K}
\label{sec:csr}
For evaluations, we adopt existing code translation evaluation metrics in our experiments, including \textbf{Exact Match (EM)}, \textbf{BLEU}, and \textbf{CodeBLEU} (details are in Appendix~\ref{sec:code_translation_metrics}). However, all these metrics are based on surface-form matching (or with some adaptations as for CodeBLEU) and are not suitable for our \textbf{program-level} translation tasks since they cannot reliably evaluate functional correctness of translated code. Moreover, in real-world software development scenarios, developers typically ensure the functionality of code by testing and debugging upon completion, rather than writing and testing multiple versions of the code to achieve the expected functionality as measured by the existing pass@k~\cite{kulal2019spoc} metric. Meanwhile, recent research shows that LLMs such as ChatGPT demonstrate preliminary code debugging capabilities~\cite{chen2023teaching,chen2023improving}.  Hence, we propose a novel and robust evaluation metric for LLM on code translation, \textbf{Debugging Success Rate@K (DSR@K)}, by measuring whether the translated code can be compiled and executed with the same behavior as the input source code, with K rounds of debugging. \textbf{To the best of our knowledge, \textit{DSR@K} is the first metric designed to accurately reflect real-world software development scenarios.} 

\textit{DSR@K} is the percentage of the samples that successfully execute and produce the expected results among all samples. Each sample is given $K$ generation and debugging attempts by an LLM.
If the generated code successfully executes and produces the expected results with these \textit{K} rounds, the sample is marked as successful. \textit{DSR@K} is computed as $\frac{1}{N} \sum_{i=1}^{N} S(i, K)$,
where $N$ denotes the total number of samples.
If the $i^{th}$ code sample succeeds within \textit{K} attempts, then $S (i, K)$ = 1; otherwise, $S (i, K)$ = 0.  Note that DSR@0 can be used for program-level code translation evaluation for any models. In this work, we employ DSR@K to evaluate the ability of LLMs such as ChatGPT for debugging code and translating code with debugging results.

\subsection{ChatGPT for Code Translation}
\label{sec:chatgpt}
The recent LLM ChatGPT demonstrates competitive performance on language generation tasks such as summarization and machine translation~\cite{yang2023harnessing,peng2023making,gao2023humanlike}. However, ChatGPT for code translation has not been systematically explored. We study the effectiveness and potential of ChatGPT on code translation and investigate strategies to improve its performance. We use \textbf{DSR@K} as the principal evaluation metric since we focus on the practical usability of ChatGPT. We use the ChatGPT API and gpt-3.5-turbo as the default model and evaluate on the \textbf{LLMTrans} dataset for all experiments. We investigate the efficacy of prompts and hyperparameters and context in zero-shot setting, then compare one-shot versus zero-shot and study Chain-of-Thought.

\paragraph{Effect of Prompts and Hyperparameters}
Prior works show that prompts can influence the performance of ChatGPT~\cite{zhong2023chatgpt,peng2023making,jiao2023chatgpt}. We set an initial prompt ``\texttt{Translate [SL] to [TL]:[SC].}'' as the baseline, where [SL] and [TL] denote the source language and the target language respectively and [SC] denotes the source code. We also add ``Do not return anything other than the translated code.'' for each prompting strategy to require ChatGPT to return only code in order to ease code execution. We design three prompt variants. Details of the experimental settings and prompt variants are in Appendix~\ref{appendix:zero-shot}. We also investigate the effect of hyperparameters on code translation performance.

As shown in Table~\ref{table:chatgpt_strategies_performance}, implementing role assignments, clarifying usage, and polite inquiry in prompts all degrade the performance compared to the baseline prompt. These results show that the baseline with the most straightforward prompt produces the best performance, possibly because it provides clear, short, and unambiguous instructions for the task to the model. More intricate prompting strategies may introduce noise and confuse ChatGPT. The performance of polite inquiry prompt is comparable to but still worse than the baseline performance.  We speculate that the improvement from polite inquiries in prior studies~\cite{f2023awesome} may stem from their explicit and comprehensive formulations which make it easier for the model to understand the task requirements. We also observe in Table~\ref{table:chatgpt_strategies_performance} that same as prior findings, BLEU and CodeBLEU have no obvious positive correlations with the debugging success rate (DSR@0). Since the reference target code exhibits the same functionality as the source language code but their execution results could differ slightly, EM also does not correlate with DSR@0. Therefore, in subsequent experiments, we only report DSR@0. We also evaluate the CodeT5+\_220M model on LLMTrans with the Many-to-Many strategy and find that DSR@0 is 0, suggesting that CodeT5+\_220M Zero-shot is unable to generate executable translation results.

\begin{table}[t]
    \begin{adjustbox}{width=0.48\textwidth}
    \begin{tabular}{cccccc}
    \toprule
        \textbf{Strategy} & \textbf{Expt \#num} & \textbf{EM} & \textbf{BLEU} & \textbf{CodeBLEU} & \textbf{DSR@0} \\
    \midrule
   Baseline & -- & 0.29 & 10.83 & 24.46 & \textbf{48.57\%}\\
   \hdashline
    \multirow{4}{*}{Role assignments}& 1 & 0.00 & \textbf{11.06} & 24.36 & 43.43\% \\
     & 2 & 0.00 & \textbf{11.06} & 24.48 & 43.14\% \\
     & 3 & 0.00 & 10.70 & 24.08 & 41.71\% \\
     & 4 & 0.00 & 10.73 & 24.08 & 40.86\% \\
\hdashline
\multirow{2}{*}{Polite inquiry} & 1 & 0.29 & 10.83 & 24.37 & \textbf{47.71\%}\\
 & 2 & \textbf{0.86} & 10.87 & 24.26 & \textbf{47.71\%} \\
\hdashline
 Clarify usage & -- & 0.29 & 10.63 & 24.11 & 44.00\%\\
\cmidrule(lr){1-6}
 Divide-and-Conquer & -- & 0.00 &  7.44 &  \textbf{25.30} & 22.86\% \\
    \bottomrule
  \end{tabular}
  \end{adjustbox}
    \caption{Zero-shot performance of ChatGPT with different prompt variants and contextual strategies. Baseline denotes ChatGPT with the baseline prompt. Details of the prompt variants (Expt \#num) are in Appendix~\ref{appendix:zero-shot}.}
  \label{table:chatgpt_strategies_performance}
\end{table}

\afterpage{
\begin{figure}[!h]
        \centering{%
        \begin{adjustbox}{width=0.4\textwidth}
        \begin{tabular}{ccccc}
          \toprule
          \textbf{$K^{th}$ Debug} & \textbf{DSR} & & \textbf{$K^{th}$ Debug} & \textbf{DSR}\\
          \midrule
          0 & 48.57\% & & 2 & 52.29\% \\
         1 & 51.43\%  & & 3 &\textbf{52.57\%} \\
          \bottomrule
        \end{tabular}
        \end{adjustbox}
        }
\captionof{table}{ChatGPT performance at the $K^{th}$ debugging.}  
        \label{table:chatgpt_debug}
\end{figure}
}

ChatGPT selects the token with the highest probability during generation. The hyperparameter \textit{temperature} influences the randomness of the generated text, while \textit{top\_p} controls the range of vocabulary considered during generation. Higher temperature or top\_p could increase diversity in the generated results from ChatGPT.  However, as shown in Table~\ref{table:chatgpt_parameters} in Appendix, independently varying temperature or top\_p does not notably change the performance of ChatGPT; hence for the other ChatGPT experiments, we set both temperature and top\_p as 0 to ensure stability an reproducibility.

\paragraph{Effect of Context} We explore a \textit{Divide-and-Conquer} strategy, which segments the source language code into snippets (e.g., functions and sub-functions), translate each snippet independently, then merge their outputs as the final result. As shown in Table~\ref{table:chatgpt_strategies_performance}, Divide-and-Conquer significantly degrades the performance. We hypothesize that lack of the global context in Divide-and-Conquer could prevent ChatGPT from considering the overall structure and variable configurations of the code for translation.

\paragraph{Effect of Self-debugging} \label{sec:self_debugging}
Since ChatGPT has shown preliminary capability in error detection and correction during code generation~\cite{shinn2023reflexion,chen2023teaching,kim2023language,nair2023dera,madaan2023selfrefine}, we use ChatGPT to perform multiple rounds of self-debugging and investigate the impact on DSR.  Specifically, ChatGPT first translates the source language code into the target language (which is Python as in our AutoTransExecuter) and then attempts to execute the translated code. If the execution passes and executing the translated code exhibits the same functionality as the source code, it is regarded as a successful execution. Otherwise, feedback from the compiler will be also fed to ChatGPT for the next round of translation, and this process is repeated until reaching a pre-defined number $K$ of debugging rounds. The whole process is shown in Table~\ref{tab:debug@k_prompts} in Appendix. As shown in Table~\ref{table:chatgpt_debug}, DSR improves significantly with multiple rounds of self-debugging. The first self-debugging improves DSR by \textbf{3\%} absolutely. Each subsequent round of self-debugging brings further gain but DSR begins to plateau after the second debugging round. This suggests that ChatGPT has limitations in its capacity to rectify errors after multiple debugging cycles, which is consistent with human behaviors.

\paragraph{Effect of One-shot}
In-context learning~\cite{brown2020language} allows the model to learn from input examples, enabling it to understand and manage each new task. This method has been validated as an effective strategy for enhancing the performance of model inference~\cite{peng2023making,liu2021pretrain}. Therefore, we explore one-shot learning for ChatGPT on code translation.
We investigate three one-shot learning sample selection strategies. Descriptions of the strategies and the corresponding prompts are in Appendix~\ref{appendix:one_shot}.

Table~\ref{table:chatgpt_strategies_performance_oneshot} shows that  all three One-shot learning strategies effectively improve DSR@0 of ChatGPT over the Zero-shot baseline. The Experiment\#2 strategy (provided contextual example has both same source and target languages as the original task) achieves the best performance, yielding \textbf{1.72\%} absolute gain in DSR@0, with Experiment \#1 (example has the same target language but different source language) and \#3 (example has different source and target languages) following closely with 1.14\% and 0.29\% absolute gains, respectively. These results show that One-shot learning entirely tailored to the translation requirements is most effective in boosting code translation performance for ChatGPT. The results corroborate previous findings in natural language translation~\cite{peng2023making} that the performance of ChatGPT is sensitive to the provided contextual example in One-shot learning.

\begin{table}
    \begin{adjustbox}{width=0.48\textwidth}
    \begin{tabular}{ccc|ccc}
    \toprule
        \textbf{Strategy} & \textbf{Expts \#num} & \textbf{DSR@0} & \textbf{Strategy} & \textbf{Expts \#num} & \textbf{DSR@0} \\
    \midrule
    Baseline & -- & \textbf{48.57\%} & \multirow{4}{*}{CoT} & 1 & 46.00\%\\
    \cmidrule(lr){1-3}
    \multirow{3}{*}{One-shot} & 1 & 49.71\% & & 2 & 42.57\%  \\
    & 2 & \textbf{50.29\%} & & 3 & \textbf{48.29\%}\\
    & 3 & 48.86\% & & 4 &  45.43\%\\
    \bottomrule
  \end{tabular}
  \end{adjustbox}
    \caption{Performance of ChatGPT with One-shot and CoT strategies compared to the Zero-shot Baseline. Details of Expt \#num are in Appendix~\ref{appendix:one_shot} and \ref{appendix:cot}.}
  \label{table:chatgpt_strategies_performance_oneshot}
\end{table}

\paragraph{Effect of Chain-of-Thought}
Chain-of-Thought (CoT) allows the model to simulate an orderly and structured way of thinking by sorting out the thinking process. It helps guide the model to output the final answer step by step~\cite{wei2023chainofthought,peng2023making,kojima2023large}. For code translation, we investigate four CoT strategies. Detailed descriptions and translation prompts for each strategy are in Appendix~\ref{appendix:cot}. As shown in Table~\ref{table:chatgpt_strategies_performance_oneshot}, CoT degrades executability of the translated code. In Experiment \#2, DSR@0 even declines by 6\% absolutely. 
We study the translation results of ChatGPT and find that when CoT strategies are applied, the model tends to translate the source code line by line, neglecting compatibility issues between libraries and functions in different languages. CoT also compromises the global planning ability of the model. These observations are consistent with the findings in ~\cite{peng2023making} that CoT may lead to word-by-word translations of natural language, thereby degrading the translation quality.

\paragraph{Fuzzy Execution} To address the limitations of existing evaluation metrics and our AutoTransExecuter, we propose another novel code translation evaluation metric \textbf{fuzzy execution} using LLMs in Section~\hyperref[sec:limitations]{Limitations}, inspired by recent progress in using LLMs as evaluation metrics for NLP tasks. Our preliminary studies evaluates the performance of ChatGPT for predicting whether a given code can be executed or not, and if executable, also for predicting the executed output. Experimental results show that using ChatGPT for fuzzy execution is not yet practical and demands future research.

\section{Conclusion}
We construct CodeTransOcean, a comprehensive code translation benchmark that includes multilingual and cross-framework datasets. We demonstrate that multilingual modeling has remarkable potential in enhancing code translation quality. We also reveal the superior code translation capability of ChatGPT and advanced strategies lead to significant performance gains. Moreover, we introduce fuzzy execution that may overcome limitations of existing metrics but requires future research. In summary, we provide a comprehensive suite of resources, tools, and baselines for code translation.

\section{Limitations}
\label{sec:limitations}

Existing match-based evaluation metrics for code translation~\cite{papineni2002bleu,ren2020codebleu,eghbali2022crystalbleu,zhou2023codebertscore,tran2019does} focus solely on semantics, overlooking executability of the code and the functional equivalence under different implementations. Execution-based metrics~\cite{kulal2019spoc,hao2022aixbench,hendrycks2021measuring,lachaux2020unsupervised,dong2023codescore} that require providing test cases are expensive to conduct in practice, and the significant overhead of executing numerous test cases and the heightened security risks during the execution process remain unresolved. It is crucial to establish an evaluation metric that overcomes these limitations. 

Our proposed DSR@K and the automated AutoTransExecuter aim to measure the executability of the code and reflect the real-world software development scenarios. However, AutoTransExecuter currently only supports Python as the target language. This is mainly due to the fact that different programming languages necessitate distinct run-time environments and libraries, making it particularly challenging to automatically detect and install the required dependencies for each code. While certain existing tools, such as Dynatrace\footnote{\url{https://www.dynatrace.com/platform/artificial-intelligence/dependency-detection/}}, can carry out dependency detection, the range of supported programming languages remains limited. Moreover, the configuration methods for compilers vary substantially among different programming languages, which further complicates automated configuration. In addition, fully automated execution systems could be exploited by malicious code, thus necessitating further security measures. Therefore, achieving this goal requires overcoming many technical and practical difficulties.

To address limitations of existing evaluation metrics and limitations of AutoTransExecuter, we propose another novel code translation evaluation metric \textbf{fuzzy execution}.

Recent studies have begun to utilize LLMs as evaluation metrics in the field of NLP~\cite{chen2023exploring,wang2023chatgpt,fu2023gptscore,kocmi2023large,ji2023exploring}. Inspired by these works, we create a new dataset \textbf{ExecuteStatus} by randomly selecting 300 executable samples from MultilingualTrans and 300 non-executable samples from the translation results of ChatGPT. Each entry in this dataset includes the execution status and, if executable, the result of the execution. We use ExecuteStatus and AutoTransExecuter to evaluate the performance of ChatGPT for predicting whether a given code can be executed or not, and if executable, also predict the executed output. The Zero-shot prompts are shown in Table~\ref{tab:fuzzy_prompts} in Appendix. For the Few-shot strategy, in addition to the Zero-shot baseline, we include an example of executable code and an example of non-executable code, as detailed in Table~\ref{tab:fuzzy_prompts}.

\begin{table}[t]
    \centering
    \begin{adjustbox}{width=0.48\textwidth}
    \begin{tabular}{cccccccccc}
      \toprule
      \multicolumn{5}{c}{\textbf{Zero-Shot}} & \multicolumn{5}{c}{\textbf{Few-Shot}} \\
      \cmidrule(lr){1-5} \cmidrule(lr){6-10}
      \multicolumn{2}{c}{\textbf{TN}} & \textbf{FP} & \textbf{FN} & \textbf{TP} & \multicolumn{2}{c}{\textbf{TN}} & \textbf{FP} & \textbf{FN} & \textbf{TP}\\
      \cmidrule(lr){1-5} \cmidrule(lr){6-10}
      \multicolumn{2}{c}{292}  & 8  & 238  & 62 & \multicolumn{2}{c}{294} & 4  & 242 & 58 \\
      \cmidrule(lr){1-2}   \cmidrule(lr){6-7}
      $\color{red}\checkmark$12 & $\color{red}\times$280  &  &  &  & $\color{red}\checkmark$14 & $\color{red}\times$282 &  & &  \\
      \bottomrule
    \end{tabular}
     \end{adjustbox}
\caption{Confusion matrix of fuzzy execution prediction by ChatGPT with Zero-shot and Few-shot settings.}
    \label{table:fuzzy_execution}
\end{table}

\begin{table}[t]
\renewcommand{\arraystretch}{2}
\centering
\begin{adjustbox}{width=0.48\textwidth}
\begin{tabular}{cccc}
\midrule
\textbf{Metrics} & \textbf{Calculation formula} & \textbf{Zero-Shot} & \textbf{Few-Shot}  \\ \midrule
\textbf{Accuracy}    & $\dfrac{\text{TP}+ \text{TN}}{\text{TP} + \text{TN} + \text{FN} + \text{FP}}$   & 59.00\%     & 59.00\%  \\ 
\textbf{Precision} & $\dfrac{\text{TP}}{\text{TP} + \text{FP}}$  & 88.57\% & 93.55\% \\ 
\textbf{Recall}      & $\dfrac{\text{T P}}{\text{TP} + \text{FN}}$ & 20.67\%  & 19.33\% \\ 
\textbf{F1 scores}     & $2 \cdot \dfrac{\text{Precision} \cdot \text{Recall}}{\text{Precision} + \text{Recall}}$ & 33.52\%  & 32.04\%  \\  \midrule
\end{tabular}
\end{adjustbox}
\caption{Performance of ChatGPT on predicting fuzzy execution.}
\label{tab:des_fuzzy_execution}
\end{table}

We define fuzzy execution as first testing the consistency between the actual pass rate and the predicted pass rate of ChatGPT, followed by further testing the accuracy in predicting execution results using ChatGPT without relying on a compiler. Since we are interested in the ability of ChatGPT to identify samples that cannot actually be executed accurately, we present the confusion matrix in Table~\ref{table:fuzzy_execution} based on the results. To evaluate the performance of ChatGPT on the fuzzy execution prediction task, we use the standard accuracy, precision, recall, and F1 scores. Experimental results based on these evaluation metrics are in Table~\ref{tab:des_fuzzy_execution}. The low accuracy, recall and F1 scores show that ChatGPT still has difficulty in identifying errors in the code, exhibiting about an 88\% tendency to predict that the code is executable. Overall, ChatGPT has low accuracy in the binary classification task of ``whether it can be executed'', and its ability to predict execution results, being at a scant 4\%, clearly requires further enhancement. Thus, using ChatGPT for fuzzy execution is not yet practical~\cite{liu2023evaluate}. Despite this, fuzzy execution with LLMs holds the potential to overcome the deficiencies of current code translation evaluation metrics. We will continue this exploration in future work.



\bibliography{emnlp2023}

\begin{thebibliography}{66}
\expandafter\ifx\csname natexlab\endcsname\relax\def\natexlab#1{#1}\fi

\bibitem[{Aharoni et~al.(2019)Aharoni, Johnson, and Firat}]{aharoni2019massively}
Roee Aharoni, Melvin Johnson, and Orhan Firat. 2019.
\newblock \href {https://doi.org/10.18653/v1/n19-1388} {Massively multilingual neural machine translation}.
\newblock In \emph{Proceedings of the 2019 Conference of the North American Chapter of the Association for Computational Linguistics: Human Language Technologies, {NAACL-HLT} 2019, Minneapolis, MN, USA, June 2-7, 2019, Volume 1 (Long and Short Papers)}, pages 3874--3884. Association for Computational Linguistics.

\bibitem[{Ahmad et~al.(2021)Ahmad, Chakraborty, Ray, and Chang}]{ahmad2021unified}
Wasi~Uddin Ahmad, Saikat Chakraborty, Baishakhi Ray, and Kai{-}Wei Chang. 2021.
\newblock \href {https://doi.org/10.18653/v1/2021.naacl-main.211} {Unified pre-training for program understanding and generation}.
\newblock In \emph{Proceedings of the 2021 Conference of the North American Chapter of the Association for Computational Linguistics: Human Language Technologies, {NAACL-HLT} 2021, Online, June 6-11, 2021}, pages 2655--2668. Association for Computational Linguistics.

\bibitem[{Ahmad et~al.(2023)Ahmad, Tushar, Chakraborty, and Chang}]{ahmad2021avatar}
Wasi~Uddin Ahmad, Md~Golam~Rahman Tushar, Saikat Chakraborty, and Kai{-}Wei Chang. 2023.
\newblock \href {https://doi.org/10.18653/v1/2023.findings-acl.143} {{AVATAR:} {A} parallel corpus for java-python program translation}.
\newblock In \emph{Findings of the Association for Computational Linguistics: {ACL} 2023, Toronto, Canada, July 9-14, 2023}, pages 2268--2281. Association for Computational Linguistics.

\bibitem[{Akın(2023)}]{f2023awesome}
Fatih~Kadir Akın. 2023.
\newblock \href {https://github.com/f/awesome-chatgpt-prompts} {awesome-chatgpt-prompts}.

\bibitem[{AlKhamissi et~al.(2023)AlKhamissi, Verma, Yu, Jin, Celikyilmaz, and Diab}]{alkhamissi2023optr}
Badr AlKhamissi, Siddharth Verma, Ping Yu, Zhijing Jin, Asli Celikyilmaz, and Mona~T. Diab. 2023.
\newblock \href {https://doi.org/10.48550/arXiv.2305.12001} {{OPT-R:} exploring the role of explanations in finetuning and prompting for reasoning skills of large language models}.
\newblock \emph{CoRR}, abs/2305.12001.

\bibitem[{Ananthaswamy(2023)}]{anil2023bigger}
Anil Ananthaswamy. 2023.
\newblock In ai, is bigger always better?
\newblock \emph{Nature}.

\bibitem[{Anil et~al.(2023)Anil, Dai, Firat, Johnson, Lepikhin, Passos, Shakeri, Taropa, Bailey, Chen, Chu, Clark, Shafey, Huang, Meier{-}Hellstern, Mishra, Moreira, Omernick, Robinson, Ruder, Tay, Xiao, Xu, Zhang, {\'{A}}brego, Ahn, Austin, Barham, Botha, Bradbury, Brahma, Brooks, Catasta, Cheng, Cherry, Choquette{-}Choo, Chowdhery, Crepy, Dave, Dehghani, Dev, Devlin, D{\'{\i}}az, Du, Dyer, Feinberg, Feng, Fienber, Freitag, Garcia, Gehrmann, Gonzalez, and et~al.}]{anil2023palm}
Rohan Anil, Andrew~M. Dai, Orhan Firat, Melvin Johnson, Dmitry Lepikhin, Alexandre Passos, Siamak Shakeri, Emanuel Taropa, Paige Bailey, Zhifeng Chen, Eric Chu, Jonathan~H. Clark, Laurent~El Shafey, Yanping Huang, Kathy Meier{-}Hellstern, Gaurav Mishra, Erica Moreira, Mark Omernick, Kevin Robinson, Sebastian Ruder, Yi~Tay, Kefan Xiao, Yuanzhong Xu, Yujing Zhang, Gustavo~Hern{\'{a}}ndez {\'{A}}brego, Junwhan Ahn, Jacob Austin, Paul Barham, Jan~A. Botha, James Bradbury, Siddhartha Brahma, Kevin Brooks, Michele Catasta, Yong Cheng, Colin Cherry, Christopher~A. Choquette{-}Choo, Aakanksha Chowdhery, Cl{\'{e}}ment Crepy, Shachi Dave, Mostafa Dehghani, Sunipa Dev, Jacob Devlin, Mark D{\'{\i}}az, Nan Du, Ethan Dyer, Vladimir Feinberg, Fangxiaoyu Feng, Vlad Fienber, Markus Freitag, Xavier Garcia, Sebastian Gehrmann, Lucas Gonzalez, and et~al. 2023.
\newblock \href {https://doi.org/10.48550/arXiv.2305.10403} {Palm 2 technical report}.
\newblock \emph{CoRR}, abs/2305.10403.

\bibitem[{Brown et~al.(2020)Brown, Mann, Ryder, Subbiah, Kaplan, Dhariwal, Neelakantan, Shyam, Sastry, Askell, Agarwal, Herbert{-}Voss, Krueger, Henighan, Child, Ramesh, Ziegler, Wu, Winter, Hesse, Chen, Sigler, Litwin, Gray, Chess, Clark, Berner, McCandlish, Radford, Sutskever, and Amodei}]{brown2020language}
Tom~B. Brown, Benjamin Mann, Nick Ryder, Melanie Subbiah, Jared Kaplan, Prafulla Dhariwal, Arvind Neelakantan, Pranav Shyam, Girish Sastry, Amanda Askell, Sandhini Agarwal, Ariel Herbert{-}Voss, Gretchen Krueger, Tom Henighan, Rewon Child, Aditya Ramesh, Daniel~M. Ziegler, Jeffrey Wu, Clemens Winter, Christopher Hesse, Mark Chen, Eric Sigler, Mateusz Litwin, Scott Gray, Benjamin Chess, Jack Clark, Christopher Berner, Sam McCandlish, Alec Radford, Ilya Sutskever, and Dario Amodei. 2020.
\newblock \href {https://proceedings.neurips.cc/paper/2020/hash/1457c0d6bfcb4967418bfb8ac142f64a-Abstract.html} {Language models are few-shot learners}.
\newblock In \emph{Advances in Neural Information Processing Systems 33: Annual Conference on Neural Information Processing Systems 2020, NeurIPS 2020, December 6-12, 2020, virtual}.

\bibitem[{Chakraborty et~al.(2022)Chakraborty, Ahmed, Ding, Devanbu, and Ray}]{chakraborty2022natgen}
Saikat Chakraborty, Toufique Ahmed, Yangruibo Ding, Premkumar~T. Devanbu, and Baishakhi Ray. 2022.
\newblock \href {https://doi.org/10.1145/3540250.3549162} {Natgen: generative pre-training by "naturalizing" source code}.
\newblock In \emph{Proceedings of the 30th {ACM} Joint European Software Engineering Conference and Symposium on the Foundations of Software Engineering, {ESEC/FSE} 2022, Singapore, Singapore, November 14-18, 2022}, pages 18--30. {ACM}.

\bibitem[{Chen et~al.(2023{\natexlab{a}})Chen, Scheurer, Korbak, Campos, Chan, Bowman, Cho, and Perez}]{chen2023improving}
Angelica Chen, J{\'{e}}r{\'{e}}my Scheurer, Tomasz Korbak, Jon~Ander Campos, Jun~Shern Chan, Samuel~R. Bowman, Kyunghyun Cho, and Ethan Perez. 2023{\natexlab{a}}.
\newblock \href {https://doi.org/10.48550/arXiv.2303.16749} {Improving code generation by training with natural language feedback}.
\newblock \emph{CoRR}, abs/2303.16749.

\bibitem[{Chen et~al.(2023{\natexlab{b}})Chen, Lin, Sch{\"{a}}rli, and Zhou}]{chen2023teaching}
Xinyun Chen, Maxwell Lin, Nathanael Sch{\"{a}}rli, and Denny Zhou. 2023{\natexlab{b}}.
\newblock \href {https://doi.org/10.48550/arXiv.2304.05128} {Teaching large language models to self-debug}.
\newblock \emph{CoRR}, abs/2304.05128.

\bibitem[{Chen et~al.(2018)Chen, Liu, and Song}]{chen2018tree}
Xinyun Chen, Chang Liu, and Dawn Song. 2018.
\newblock \href {https://proceedings.neurips.cc/paper/2018/hash/d759175de8ea5b1d9a2660e45554894f-Abstract.html} {Tree-to-tree neural networks for program translation}.
\newblock In \emph{Advances in Neural Information Processing Systems 31: Annual Conference on Neural Information Processing Systems 2018, NeurIPS 2018, December 3-8, 2018, Montr{\'{e}}al, Canada}, pages 2552--2562.

\bibitem[{Chen et~al.(2023{\natexlab{c}})Chen, Wang, Jiang, Shi, and Xu}]{chen2023exploring}
Yi~Chen, Rui Wang, Haiyun Jiang, Shuming Shi, and Ruifeng Xu. 2023{\natexlab{c}}.
\newblock \href {https://doi.org/10.48550/arXiv.2304.00723} {Exploring the use of large language models for reference-free text quality evaluation: {A} preliminary empirical study}.
\newblock \emph{CoRR}, abs/2304.00723.

\bibitem[{Dong et~al.(2023)Dong, Ding, Jiang, Li, Li, and Jin}]{dong2023codescore}
Yihong Dong, Jiazheng Ding, Xue Jiang, Zhuo Li, Ge~Li, and Zhi Jin. 2023.
\newblock \href {https://doi.org/10.48550/arXiv.2301.09043} {Codescore: Evaluating code generation by learning code execution}.
\newblock \emph{CoRR}, abs/2301.09043.

\bibitem[{Eghbali and Pradel(2022)}]{eghbali2022crystalbleu}
Aryaz Eghbali and Michael Pradel. 2022.
\newblock \href {https://doi.org/10.1145/3551349.3556903} {Crystalbleu: Precisely and efficiently measuring the similarity of code}.
\newblock In \emph{37th {IEEE/ACM} International Conference on Automated Software Engineering, {ASE} 2022, Rochester, MI, USA, October 10-14, 2022}, pages 28:1--28:12. {ACM}.

\bibitem[{Feng et~al.(2020)Feng, Guo, Tang, Duan, Feng, Gong, Shou, Qin, Liu, Jiang, and Zhou}]{feng2020codebert}
Zhangyin Feng, Daya Guo, Duyu Tang, Nan Duan, Xiaocheng Feng, Ming Gong, Linjun Shou, Bing Qin, Ting Liu, Daxin Jiang, and Ming Zhou. 2020.
\newblock \href {https://doi.org/10.18653/v1/2020.findings-emnlp.139} {Codebert: {A} pre-trained model for programming and natural languages}.
\newblock In \emph{Findings of the Association for Computational Linguistics: {EMNLP} 2020, Online Event, 16-20 November 2020}, volume {EMNLP} 2020 of \emph{Findings of {ACL}}, pages 1536--1547. Association for Computational Linguistics.

\bibitem[{Fu et~al.(2023)Fu, Ng, Jiang, and Liu}]{fu2023gptscore}
Jinlan Fu, See{-}Kiong Ng, Zhengbao Jiang, and Pengfei Liu. 2023.
\newblock \href {https://doi.org/10.48550/arXiv.2302.04166} {Gptscore: Evaluate as you desire}.
\newblock \emph{CoRR}, abs/2302.04166.

\bibitem[{Gao et~al.(2023)Gao, Ruan, Sun, Yin, Yang, and Wan}]{gao2023humanlike}
Mingqi Gao, Jie Ruan, Renliang Sun, Xunjian Yin, Shiping Yang, and Xiaojun Wan. 2023.
\newblock \href {https://doi.org/10.48550/arXiv.2304.02554} {Human-like summarization evaluation with chatgpt}.
\newblock \emph{CoRR}, abs/2304.02554.

\bibitem[{Georgiou et~al.(2022)Georgiou, Kechagia, Sharma, Sarro, and Zou}]{georgiou2022green}
Stefanos Georgiou, Maria Kechagia, Tushar Sharma, Federica Sarro, and Ying Zou. 2022.
\newblock \href {https://doi.org/10.1145/3510003.3510221} {Green {AI:} do deep learning frameworks have different costs?}
\newblock In \emph{44th {IEEE/ACM} 44th International Conference on Software Engineering, {ICSE} 2022, Pittsburgh, PA, USA, May 25-27, 2022}, pages 1082--1094. {ACM}.

\bibitem[{Guo et~al.(2021)Guo, Ren, Lu, Feng, Tang, Liu, Zhou, Duan, Svyatkovskiy, Fu, Tufano, Deng, Clement, Drain, Sundaresan, Yin, Jiang, and Zhou}]{guo2020graphcodebert}
Daya Guo, Shuo Ren, Shuai Lu, Zhangyin Feng, Duyu Tang, Shujie Liu, Long Zhou, Nan Duan, Alexey Svyatkovskiy, Shengyu Fu, Michele Tufano, Shao~Kun Deng, Colin~B. Clement, Dawn Drain, Neel Sundaresan, Jian Yin, Daxin Jiang, and Ming Zhou. 2021.
\newblock \href {https://openreview.net/forum?id=jLoC4ez43PZ} {Graphcodebert: Pre-training code representations with data flow}.
\newblock In \emph{9th International Conference on Learning Representations, {ICLR} 2021, Virtual Event, Austria, May 3-7, 2021}. OpenReview.net.

\bibitem[{Hao et~al.(2022)Hao, Li, Liu, Miao, Zong, Jiang, Liu, and Wei}]{hao2022aixbench}
Yiyang Hao, Ge~Li, Yongqiang Liu, Xiaowei Miao, He~Zong, Siyuan Jiang, Yang Liu, and He~Wei. 2022.
\newblock \href {https://doi.org/10.48550/arXiv.2206.13179} {Aixbench: {A} code generation benchmark dataset}.
\newblock \emph{CoRR}, abs/2206.13179.

\bibitem[{Hendrycks et~al.(2021)Hendrycks, Basart, Kadavath, Mazeika, Arora, Guo, Burns, Puranik, He, Song, and Steinhardt}]{hendrycks2021measuring}
Dan Hendrycks, Steven Basart, Saurav Kadavath, Mantas Mazeika, Akul Arora, Ethan Guo, Collin Burns, Samir Puranik, Horace He, Dawn Song, and Jacob Steinhardt. 2021.
\newblock \href {https://datasets-benchmarks-proceedings.neurips.cc/paper/2021/hash/c24cd76e1ce41366a4bbe8a49b02a028-Abstract-round2.html} {Measuring coding challenge competence with {APPS}}.
\newblock In \emph{Proceedings of the Neural Information Processing Systems Track on Datasets and Benchmarks 1, NeurIPS Datasets and Benchmarks 2021, December 2021, virtual}.

\bibitem[{Ji et~al.(2023)Ji, Gong, Peng, Ni, Sun, Pan, Ma, and Li}]{ji2023exploring}
Yunjie Ji, Yan Gong, Yiping Peng, Chao Ni, Peiyan Sun, Dongyu Pan, Baochang Ma, and Xiangang Li. 2023.
\newblock \href {https://doi.org/10.48550/arXiv.2303.07610} {Exploring chatgpt's ability to rank content: {A} preliminary study on consistency with human preferences}.
\newblock \emph{CoRR}, abs/2303.07610.

\bibitem[{Jiao et~al.(2023)Jiao, Wang, tse Huang, Wang, and Tu}]{jiao2023chatgpt}
Wenxiang Jiao, Wenxuan Wang, Jen tse Huang, Xing Wang, and Zhaopeng Tu. 2023.
\newblock \href {http://arxiv.org/abs/2301.08745} {Is chatgpt a good translator? yes with gpt-4 as the engine}.

\bibitem[{Johnson et~al.(2017)Johnson, Schuster, Le, Krikun, Wu, Chen, Thorat, Vi{\'{e}}gas, Wattenberg, Corrado, Hughes, and Dean}]{johnson2017googles}
Melvin Johnson, Mike Schuster, Quoc~V. Le, Maxim Krikun, Yonghui Wu, Zhifeng Chen, Nikhil Thorat, Fernanda~B. Vi{\'{e}}gas, Martin Wattenberg, Greg Corrado, Macduff Hughes, and Jeffrey Dean. 2017.
\newblock \href {https://doi.org/10.1162/tacl\_a\_00065} {Google's multilingual neural machine translation system: Enabling zero-shot translation}.
\newblock \emph{Trans. Assoc. Comput. Linguistics}, 5:339--351.

\bibitem[{Kim et~al.(2023)Kim, Baldi, and McAleer}]{kim2023language}
Geunwoo Kim, Pierre Baldi, and Stephen McAleer. 2023.
\newblock \href {https://doi.org/10.48550/arXiv.2303.17491} {Language models can solve computer tasks}.
\newblock \emph{CoRR}, abs/2303.17491.

\bibitem[{Kocmi and Federmann(2023)}]{kocmi2023large}
Tom Kocmi and Christian Federmann. 2023.
\newblock \href {https://aclanthology.org/2023.eamt-1.19} {Large language models are state-of-the-art evaluators of translation quality}.
\newblock In \emph{Proceedings of the 24th Annual Conference of the European Association for Machine Translation, {EAMT} 2023, Tampere, Finland, 12-15 June 2023}, pages 193--203. European Association for Machine Translation.

\bibitem[{Kojima et~al.(2022)Kojima, Gu, Reid, Matsuo, and Iwasawa}]{kojima2023large}
Takeshi Kojima, Shixiang~Shane Gu, Machel Reid, Yutaka Matsuo, and Yusuke Iwasawa. 2022.
\newblock \href {http://papers.nips.cc/paper\_files/paper/2022/hash/8bb0d291acd4acf06ef112099c16f326-Abstract-Conference.html} {Large language models are zero-shot reasoners}.
\newblock In \emph{NeurIPS}.

\bibitem[{Kulal et~al.(2019)Kulal, Pasupat, Chandra, Lee, Padon, Aiken, and Liang}]{kulal2019spoc}
Sumith Kulal, Panupong Pasupat, Kartik Chandra, Mina Lee, Oded Padon, Alex Aiken, and Percy Liang. 2019.
\newblock \href {https://proceedings.neurips.cc/paper/2019/hash/7298332f04ac004a0ca44cc69ecf6f6b-Abstract.html} {Spoc: Search-based pseudocode to code}.
\newblock In \emph{Advances in Neural Information Processing Systems 32: Annual Conference on Neural Information Processing Systems 2019, NeurIPS 2019, December 8-14, 2019, Vancouver, BC, Canada}, pages 11883--11894.

\bibitem[{Liu et~al.(2023{\natexlab{a}})Liu, Yuan, Fu, Jiang, Hayashi, and Neubig}]{liu2021pretrain}
Pengfei Liu, Weizhe Yuan, Jinlan Fu, Zhengbao Jiang, Hiroaki Hayashi, and Graham Neubig. 2023{\natexlab{a}}.
\newblock \href {https://doi.org/10.1145/3560815} {Pre-train, prompt, and predict: {A} systematic survey of prompting methods in natural language processing}.
\newblock \emph{{ACM} Comput. Surv.}, 55(9):195:1--195:35.

\bibitem[{Liu et~al.(2019)Liu, Ott, Goyal, Du, Joshi, Chen, Levy, Lewis, Zettlemoyer, and Stoyanov}]{DBLP:journals/corr/abs-1907-11692}
Yinhan Liu, Myle Ott, Naman Goyal, Jingfei Du, Mandar Joshi, Danqi Chen, Omer Levy, Mike Lewis, Luke Zettlemoyer, and Veselin Stoyanov. 2019.
\newblock \href {http://arxiv.org/abs/1907.11692} {Roberta: {A} robustly optimized {BERT} pretraining approach}.
\newblock \emph{CoRR}, abs/1907.11692.

\bibitem[{Liu et~al.(2023{\natexlab{b}})Liu, Feng, Wang, Zhang, and Sch{\"{u}}tze}]{liu2023evaluate}
Yongkang Liu, Shi Feng, Daling Wang, Yifei Zhang, and Hinrich Sch{\"{u}}tze. 2023{\natexlab{b}}.
\newblock \href {https://doi.org/10.48550/arXiv.2305.14658} {Evaluate what you can't evaluate: Unassessable generated responses quality}.
\newblock \emph{CoRR}, abs/2305.14658.

\bibitem[{Lu et~al.(2021)Lu, Guo, Ren, Huang, Svyatkovskiy, Blanco, Clement, Drain, Jiang, Tang, Li, Zhou, Shou, Zhou, Tufano, Gong, Zhou, Duan, Sundaresan, Deng, Fu, and Liu}]{lu2021codexglue}
Shuai Lu, Daya Guo, Shuo Ren, Junjie Huang, Alexey Svyatkovskiy, Ambrosio Blanco, Colin~B. Clement, Dawn Drain, Daxin Jiang, Duyu Tang, Ge~Li, Lidong Zhou, Linjun Shou, Long Zhou, Michele Tufano, Ming Gong, Ming Zhou, Nan Duan, Neel Sundaresan, Shao~Kun Deng, Shengyu Fu, and Shujie Liu. 2021.
\newblock \href {https://datasets-benchmarks-proceedings.neurips.cc/paper/2021/hash/c16a5320fa475530d9583c34fd356ef5-Abstract-round1.html} {Codexglue: {A} machine learning benchmark dataset for code understanding and generation}.
\newblock In \emph{Proceedings of the Neural Information Processing Systems Track on Datasets and Benchmarks 1, NeurIPS Datasets and Benchmarks 2021, December 2021, virtual}.

\bibitem[{Luo et~al.(2023)Luo, Xu, Zhao, Sun, Geng, Hu, Tao, Ma, Lin, and Jiang}]{luo2023wizardcoder}
Ziyang Luo, Can Xu, Pu~Zhao, Qingfeng Sun, Xiubo Geng, Wenxiang Hu, Chongyang Tao, Jing Ma, Qingwei Lin, and Daxin Jiang. 2023.
\newblock \href {https://doi.org/10.48550/arXiv.2306.08568} {Wizardcoder: Empowering code large language models with evol-instruct}.
\newblock \emph{CoRR}, abs/2306.08568.

\bibitem[{Madaan et~al.(2023)Madaan, Tandon, Gupta, Hallinan, Gao, Wiegreffe, Alon, Dziri, Prabhumoye, Yang, Welleck, Majumder, Gupta, Yazdanbakhsh, and Clark}]{madaan2023selfrefine}
Aman Madaan, Niket Tandon, Prakhar Gupta, Skyler Hallinan, Luyu Gao, Sarah Wiegreffe, Uri Alon, Nouha Dziri, Shrimai Prabhumoye, Yiming Yang, Sean Welleck, Bodhisattwa~Prasad Majumder, Shashank Gupta, Amir Yazdanbakhsh, and Peter Clark. 2023.
\newblock \href {https://doi.org/10.48550/arXiv.2303.17651} {Self-refine: Iterative refinement with self-feedback}.
\newblock \emph{CoRR}, abs/2303.17651.

\bibitem[{Mohammadshahi et~al.(2022)Mohammadshahi, Nikoulina, Berard, Brun, Henderson, and Besacier}]{mohammadshahi2022small100}
Alireza Mohammadshahi, Vassilina Nikoulina, Alexandre Berard, Caroline Brun, James Henderson, and Laurent Besacier. 2022.
\newblock \href {https://doi.org/10.18653/v1/2022.emnlp-main.571} {Small-100: Introducing shallow multilingual machine translation model for low-resource languages}.
\newblock In \emph{Proceedings of the 2022 Conference on Empirical Methods in Natural Language Processing, {EMNLP} 2022, Abu Dhabi, United Arab Emirates, December 7-11, 2022}, pages 8348--8359. Association for Computational Linguistics.

\bibitem[{Nair et~al.(2023)Nair, Schumacher, Tso, and Kannan}]{nair2023dera}
Varun Nair, Elliot Schumacher, Geoffrey~J. Tso, and Anitha Kannan. 2023.
\newblock \href {https://doi.org/10.48550/arXiv.2303.17071} {{DERA:} enhancing large language model completions with dialog-enabled resolving agents}.
\newblock \emph{CoRR}, abs/2303.17071.

\bibitem[{Nguyen et~al.(2013)Nguyen, Nguyen, and Nguyen}]{nguyen2013lexical}
Anh~Tuan Nguyen, Tung~Thanh Nguyen, and Tien~N. Nguyen. 2013.
\newblock \href {https://doi.org/10.1145/2491411.2494584} {Lexical statistical machine translation for language migration}.
\newblock In \emph{Joint Meeting of the European Software Engineering Conference and the {ACM} {SIGSOFT} Symposium on the Foundations of Software Engineering, ESEC/FSE'13, Saint Petersburg, Russian Federation, August 18-26, 2013}, pages 651--654. {ACM}.

\bibitem[{Nguyen and Chiang(2017)}]{nguyen2017transfer}
Toan~Q. Nguyen and David Chiang. 2017.
\newblock \href {https://aclanthology.org/I17-2050/} {Transfer learning across low-resource, related languages for neural machine translation}.
\newblock In \emph{Proceedings of the Eighth International Joint Conference on Natural Language Processing, {IJCNLP} 2017, Taipei, Taiwan, November 27 - December 1, 2017, Volume 2: Short Papers}, pages 296--301. Asian Federation of Natural Language Processing.

\bibitem[{OpenAI(2023)}]{openai2023gpt4}
OpenAI. 2023.
\newblock \href {https://doi.org/10.48550/arXiv.2303.08774} {{GPT-4} technical report}.
\newblock \emph{CoRR}, abs/2303.08774.

\bibitem[{Opidi(2020)}]{freecodecamp2020}
Alfrick Opidi. 2020.
\newblock \href {https://www.freecodecamp.org/news/legacy-software-maintenance-challenges/} {Why your legacy software is hard to maintain and what to do about it}.

\bibitem[{Papineni et~al.(2002)Papineni, Roukos, Ward, and Zhu}]{papineni2002bleu}
Kishore Papineni, Salim Roukos, Todd Ward, and Wei{-}Jing Zhu. 2002.
\newblock \href {https://doi.org/10.3115/1073083.1073135} {Bleu: a method for automatic evaluation of machine translation}.
\newblock In \emph{Proceedings of the 40th Annual Meeting of the Association for Computational Linguistics, July 6-12, 2002, Philadelphia, PA, {USA}}, pages 311--318. {ACL}.

\bibitem[{Peng et~al.(2023)Peng, Ding, Zhong, Shen, Liu, Zhang, Ouyang, and Tao}]{peng2023making}
Keqin Peng, Liang Ding, Qihuang Zhong, Li~Shen, Xuebo Liu, Min Zhang, Yuanxin Ouyang, and Dacheng Tao. 2023.
\newblock \href {https://doi.org/10.48550/arXiv.2303.13780} {Towards making the most of chatgpt for machine translation}.
\newblock \emph{CoRR}, abs/2303.13780.

\bibitem[{Puri et~al.(2021)Puri, Kung, Janssen, Zhang, Domeniconi, Zolotov, Dolby, Chen, Choudhury, Decker, Thost, Buratti, Pujar, and Finkler}]{puri2021project}
Ruchir Puri, David~S. Kung, Geert Janssen, Wei Zhang, Giacomo Domeniconi, Vladimir Zolotov, Julian Dolby, Jie Chen, Mihir~R. Choudhury, Lindsey Decker, Veronika Thost, Luca Buratti, Saurabh Pujar, and Ulrich Finkler. 2021.
\newblock \href {http://arxiv.org/abs/2105.12655} {Project codenet: {A} large-scale {AI} for code dataset for learning a diversity of coding tasks}.
\newblock \emph{CoRR}, abs/2105.12655.

\bibitem[{Ren et~al.(2020)Ren, Guo, Lu, Zhou, Liu, Tang, Sundaresan, Zhou, Blanco, and Ma}]{ren2020codebleu}
Shuo Ren, Daya Guo, Shuai Lu, Long Zhou, Shujie Liu, Duyu Tang, Neel Sundaresan, Ming Zhou, Ambrosio Blanco, and Shuai Ma. 2020.
\newblock \href {http://arxiv.org/abs/2009.10297} {Codebleu: a method for automatic evaluation of code synthesis}.
\newblock \emph{CoRR}, abs/2009.10297.

\bibitem[{Rithy et~al.(2022)Rithy, Hossain~Shakil, Mondal, Sultana, and Shah}]{10055851}
Israt~Jahan Rithy, Hasib Hossain~Shakil, Niloy Mondal, Fatema Sultana, and Faisal~Muhammad Shah. 2022.
\newblock \href {https://doi.org/10.1109/ICCIT57492.2022.10055851} {Xtest: A parallel multilingual corpus with test cases for code translation and its evaluation*}.
\newblock In \emph{2022 25th International Conference on Computer and Information Technology (ICCIT)}, pages 623--628.

\bibitem[{Rozi{\`{e}}re et~al.(2020)Rozi{\`{e}}re, Lachaux, Chanussot, and Lample}]{lachaux2020unsupervised}
Baptiste Rozi{\`{e}}re, Marie{-}Anne Lachaux, Lowik Chanussot, and Guillaume Lample. 2020.
\newblock \href {https://proceedings.neurips.cc/paper/2020/hash/ed23fbf18c2cd35f8c7f8de44f85c08d-Abstract.html} {Unsupervised translation of programming languages}.
\newblock In \emph{Advances in Neural Information Processing Systems 33: Annual Conference on Neural Information Processing Systems 2020, NeurIPS 2020, December 6-12, 2020, virtual}.

\bibitem[{Rozi{\`{e}}re et~al.(2022)Rozi{\`{e}}re, Zhang, Charton, Harman, Synnaeve, and Lample}]{roziere2022leveraging}
Baptiste Rozi{\`{e}}re, Jie Zhang, Fran{\c{c}}ois Charton, Mark Harman, Gabriel Synnaeve, and Guillaume Lample. 2022.
\newblock \href {https://openreview.net/forum?id=cmt-6KtR4c4} {Leveraging automated unit tests for unsupervised code translation}.
\newblock In \emph{The Tenth International Conference on Learning Representations, {ICLR} 2022, Virtual Event, April 25-29, 2022}. OpenReview.net.

\bibitem[{Shinn et~al.(2023)Shinn, Cassano, Labash, Gopinath, Narasimhan, and Yao}]{shinn2023reflexion}
Noah Shinn, Federico Cassano, Beck Labash, Ashwin Gopinath, Karthik Narasimhan, and Shunyu Yao. 2023.
\newblock \href {http://arxiv.org/abs/2303.11366} {Reflexion: Language agents with verbal reinforcement learning}.

\bibitem[{Touvron et~al.(2023)Touvron, Lavril, Izacard, Martinet, Lachaux, Lacroix, Rozière, Goyal, Hambro, Azhar, Rodriguez, Joulin, Grave, and Lample}]{touvron2023llama}
Hugo Touvron, Thibaut Lavril, Gautier Izacard, Xavier Martinet, Marie-Anne Lachaux, Timothée Lacroix, Baptiste Rozière, Naman Goyal, Eric Hambro, Faisal Azhar, Aurelien Rodriguez, Armand Joulin, Edouard Grave, and Guillaume Lample. 2023.
\newblock \href {http://arxiv.org/abs/2302.13971} {Llama: Open and efficient foundation language models}.

\bibitem[{Tran et~al.(2019)Tran, Tran, Nguyen, Nguyen, and Nguyen}]{tran2019does}
Ngoc~M. Tran, Hieu Tran, Son Nguyen, Hoan Nguyen, and Tien~N. Nguyen. 2019.
\newblock \href {https://doi.org/10.1109/ICPC.2019.00034} {Does {BLEU} score work for code migration?}
\newblock In \emph{Proceedings of the 27th International Conference on Program Comprehension, {ICPC} 2019, Montreal, QC, Canada, May 25-31, 2019}, pages 165--176. {IEEE} / {ACM}.

\bibitem[{Wang et~al.(2023{\natexlab{a}})Wang, Liang, Meng, Shi, Li, Xu, Qu, and Zhou}]{wang2023chatgpt}
Jiaan Wang, Yunlong Liang, Fandong Meng, Haoxiang Shi, Zhixu Li, Jinan Xu, Jianfeng Qu, and Jie Zhou. 2023{\natexlab{a}}.
\newblock \href {https://doi.org/10.48550/arXiv.2303.04048} {Is chatgpt a good {NLG} evaluator? {A} preliminary study}.
\newblock \emph{CoRR}, abs/2303.04048.

\bibitem[{Wang et~al.(2020)Wang, Zhai, and Hassan}]{wang2020multitask}
Yiren Wang, ChengXiang Zhai, and Hany Hassan. 2020.
\newblock \href {https://doi.org/10.18653/v1/2020.emnlp-main.75} {Multi-task learning for multilingual neural machine translation}.
\newblock In \emph{Proceedings of the 2020 Conference on Empirical Methods in Natural Language Processing, {EMNLP} 2020, Online, November 16-20, 2020}, pages 1022--1034. Association for Computational Linguistics.

\bibitem[{Wang et~al.(2023{\natexlab{b}})Wang, Le, Gotmare, Bui, Li, and Hoi}]{wang2023codet5}
Yue Wang, Hung Le, Akhilesh~Deepak Gotmare, Nghi D.~Q. Bui, Junnan Li, and Steven C.~H. Hoi. 2023{\natexlab{b}}.
\newblock \href {https://doi.org/10.48550/arXiv.2305.07922} {Codet5+: Open code large language models for code understanding and generation}.
\newblock \emph{CoRR}, abs/2305.07922.

\bibitem[{Wang et~al.(2021)Wang, Wang, Joty, and Hoi}]{wang2021codet5}
Yue Wang, Weishi Wang, Shafiq~R. Joty, and Steven C.~H. Hoi. 2021.
\newblock \href {https://doi.org/10.18653/v1/2021.emnlp-main.685} {Codet5: Identifier-aware unified pre-trained encoder-decoder models for code understanding and generation}.
\newblock In \emph{Proceedings of the 2021 Conference on Empirical Methods in Natural Language Processing, {EMNLP} 2021, Virtual Event / Punta Cana, Dominican Republic, 7-11 November, 2021}, pages 8696--8708. Association for Computational Linguistics.

\bibitem[{Wei et~al.(2022)Wei, Wang, Schuurmans, Bosma, Ichter, Xia, Chi, Le, and Zhou}]{wei2023chainofthought}
Jason Wei, Xuezhi Wang, Dale Schuurmans, Maarten Bosma, Brian Ichter, Fei Xia, Ed~H. Chi, Quoc~V. Le, and Denny Zhou. 2022.
\newblock \href {http://papers.nips.cc/paper\_files/paper/2022/hash/9d5609613524ecf4f15af0f7b31abca4-Abstract-Conference.html} {Chain-of-thought prompting elicits reasoning in large language models}.
\newblock In \emph{NeurIPS}.

\bibitem[{Weisz et~al.(2021)Weisz, Muller, Houde, Richards, Ross, Martinez, Agarwal, and Talamadupula}]{weisz2021perfection}
Justin~D. Weisz, Michael~J. Muller, Stephanie Houde, John~T. Richards, Steven~I. Ross, Fernando Martinez, Mayank Agarwal, and Kartik Talamadupula. 2021.
\newblock \href {https://doi.org/10.1145/3397481.3450656} {Perfection not required? human-ai partnerships in code translation}.
\newblock In \emph{{IUI} '21: 26th International Conference on Intelligent User Interfaces, College Station, TX, USA, April 13-17, 2021}, pages 402--412. {ACM}.

\bibitem[{Weisz et~al.(2022)Weisz, Muller, Ross, Martinez, Houde, Agarwal, Talamadupula, and Richards}]{weisz2022better}
Justin~D. Weisz, Michael~J. Muller, Steven~I. Ross, Fernando Martinez, Stephanie Houde, Mayank Agarwal, Kartik Talamadupula, and John~T. Richards. 2022.
\newblock \href {https://doi.org/10.1145/3490099.3511157} {Better together? an evaluation of ai-supported code translation}.
\newblock In \emph{{IUI} 2022: 27th International Conference on Intelligent User Interfaces, Helsinki, Finland, March 22 - 25, 2022}, pages 369--391. {ACM}.

\bibitem[{Wu et~al.(2023)Wu, Gong, Shou, Liang, and Jiang}]{wu2023large}
Ning Wu, Ming Gong, Linjun Shou, Shining Liang, and Daxin Jiang. 2023.
\newblock \href {https://doi.org/10.1007/978-3-031-44693-1\_54} {Large language models are diverse role-players for summarization evaluation}.
\newblock In \emph{Natural Language Processing and Chinese Computing - 12th National {CCF} Conference, {NLPCC} 2023, Foshan, China, October 12-15, 2023, Proceedings, Part {I}}, volume 14302 of \emph{Lecture Notes in Computer Science}, pages 695--707. Springer.

\bibitem[{Yang et~al.(2023)Yang, Jin, Tang, Han, Feng, Jiang, Yin, and Hu}]{yang2023harnessing}
Jingfeng Yang, Hongye Jin, Ruixiang Tang, Xiaotian Han, Qizhang Feng, Haoming Jiang, Bing Yin, and Xia Hu. 2023.
\newblock \href {https://doi.org/10.48550/arXiv.2304.13712} {Harnessing the power of llms in practice: {A} survey on chatgpt and beyond}.
\newblock \emph{CoRR}, abs/2304.13712.

\bibitem[{Zhong et~al.(2023)Zhong, Ding, Liu, Du, and Tao}]{zhong2023chatgpt}
Qihuang Zhong, Liang Ding, Juhua Liu, Bo~Du, and Dacheng Tao. 2023.
\newblock \href {https://doi.org/10.48550/arXiv.2302.10198} {Can chatgpt understand too? {A} comparative study on chatgpt and fine-tuned {BERT}}.
\newblock \emph{CoRR}, abs/2302.10198.

\bibitem[{Zhou et~al.(2023)Zhou, Alon, Agarwal, and Neubig}]{zhou2023codebertscore}
Shuyan Zhou, Uri Alon, Sumit Agarwal, and Graham Neubig. 2023.
\newblock \href {https://doi.org/10.48550/arXiv.2302.05527} {Codebertscore: Evaluating code generation with pretrained models of code}.
\newblock \emph{CoRR}, abs/2302.05527.

\bibitem[{Zhu et~al.(2022{\natexlab{a}})Zhu, Jain, Suresh, Ravindran, Tipirneni, and Reddy}]{zhu2022xlcost}
Ming Zhu, Aneesh Jain, Karthik Suresh, Roshan Ravindran, Sindhu Tipirneni, and Chandan~K. Reddy. 2022{\natexlab{a}}.
\newblock \href {https://doi.org/10.48550/arXiv.2206.08474} {Xlcost: {A} benchmark dataset for cross-lingual code intelligence}.
\newblock \emph{CoRR}, abs/2206.08474.

\bibitem[{Zhu et~al.(2022{\natexlab{b}})Zhu, Suresh, and Reddy}]{zhu2022multilingual}
Ming Zhu, Karthik Suresh, and Chandan~K. Reddy. 2022{\natexlab{b}}.
\newblock \href {https://doi.org/10.1609/aaai.v36i10.21434} {Multilingual code snippets training for program translation}.
\newblock In \emph{Thirty-Sixth {AAAI} Conference on Artificial Intelligence, {AAAI} 2022, Thirty-Fourth Conference on Innovative Applications of Artificial Intelligence, {IAAI} 2022, The Twelveth Symposium on Educational Advances in Artificial Intelligence, {EAAI} 2022 Virtual Event, February 22 - March 1, 2022}, pages 11783--11790. {AAAI} Press.

\bibitem[{Zhu et~al.(2023)Zhu, Liu, Dong, Xu, Kong, Chen, Li, and Huang}]{zhu2023multilingual}
Wenhao Zhu, Hongyi Liu, Qingxiu Dong, Jingjing Xu, Lingpeng Kong, Jiajun Chen, Lei Li, and Shujian Huang. 2023.
\newblock \href {https://doi.org/10.48550/arXiv.2304.04675} {Multilingual machine translation with large language models: Empirical results and analysis}.
\newblock \emph{CoRR}, abs/2304.04675.

\bibitem[{Zoph et~al.(2016)Zoph, Yuret, May, and Knight}]{zoph2016transfer}
Barret Zoph, Deniz Yuret, Jonathan May, and Kevin Knight. 2016.
\newblock \href {https://doi.org/10.18653/v1/d16-1163} {Transfer learning for low-resource neural machine translation}.
\newblock In \emph{Proceedings of the 2016 Conference on Empirical Methods in Natural Language Processing, {EMNLP} 2016, Austin, Texas, USA, November 1-4, 2016}, pages 1568--1575. The Association for Computational Linguistics.

\end{thebibliography}
\bibliographystyle{acl_natbib}

\appendix

\section{Appendix}
\label{sec:appendix}

\subsection{Related Work}
\label{appendix:related_work}

\subsubsection{Code Translation Methods}
\textbf{Naive Copy} directly duplicates the source code as the target code without making any modifications. Given that the results produced by this method are often unusable, it is treated as the lower bound of performance for code translation. Early code translation relies heavily on manual rewriting, which requires developers to have a deep understanding of both source and target languages along with the ability to navigate various complex programming structures and semantic challenges. This method is inefficient, costly, and prone to errors.

Automatic code translation methods fall into several categories. \textbf{Compilers and transpilers}\footnote{\url{https://en.wikipedia.org/wiki/Source-to-source_compiler}} can automatically translate the source code into a target language, significantly saving time and effort. However, these methods cannot fully preserve all the linguistic features and behaviors of the source code, nor can they comprehend the intent and semantics inherent to the source code as humans do. \textbf{Rule-based methods} \cite{weisz2021perfection,weisz2022better,lachaux2020unsupervised} treat the code translation task as a program synthesis problem. They define a set of transformation rules and employ the rules or pattern matching for code translation. Research on rule-based methods is quite scarce, mainly because they overly rely on the completeness of the rules and also require a considerable amount of manual preprocessing. 

\textbf{Neural network based methods} have become dominant in the field of code translation in recent years. These methods mainly treat code translation as a sequence-to-sequence generation problem. Among them, Chen et al.~\cite{chen2018tree} are the first to successfully apply neural networks to code translation, designing a tree-to-tree neural model. CodeBERT~\cite{feng2020codebert} significantly improves code translation accuracy by pretraining models with masked language modeling and replaced token detection.  GraphCodeBERT~\cite{guo2020graphcodebert} further improves code translation accuracy by introducing two additional pre-training tasks as edge prediction and node alignment. CodeT5~\cite{wang2021codet5}, based on the Transformer encoder-decoder architecture, achieves excellent performance on code translation through four pre-training tasks, namely, masked span prediction, identifier tagging, masked identifier prediction, and bimodal dual generation. With a similar architecture as CodeT5, PLBART~\cite{ahmad2021unified} adopts three tasks of token masking, token deletion and token infilling for denoising seq2seq pre-training, which enables PLBART to infer language syntax and semantics and to learn how to generate language coherently. NatGen~\cite{chakraborty2022natgen} forces the model to learn to capture intent of the source code by setting up ``Code-Naturalization'' tasks during pre-training, and forces the model to make the generated code closer to the human-written style.

In the line of neural network based methods, recently released \textbf{large language models (LLMs)} (e.g., ChatGPT~\cite{openai2023gpt4}) have shown remarkable performance in a wide range of NLP tasks with instructions and a few in-context examples. ChatGPT is built upon GPT and is optimized with Reinforcement Learning from Human Feedback. ChatGPT can efficiently understand and generate code sequences, and can self-learn from human feedback to improve the quality and accuracy of its outputs. 
This significant advancement has markedly propelled progress in the field of code translation.

\afterpage{
\begin{table*}[ht]
  \centering
  \renewcommand{\arraystretch}{1}
  \begin{tabular}{ccccc|ccccc}
    \toprule
    \textbf{Method} & \textbf{Train} & \textbf{Dev} & \textbf{Test} & \textbf{Total} & \textbf{Method} & \textbf{Train} & \textbf{Dev} & \textbf{Test} & \textbf{Total}\\
    \midrule
C $\leftrightarrow$ C\# & 796 & 84 & 169 & 1049 & C $\leftrightarrow$ C++ & 799 & 149 & 298 & 1246 \\
C $\leftrightarrow$ Go & 877 & 227 & 454 & 1558 & C $\leftrightarrow$ Java & 813 & 171 & 343 & 1327 \\
C $\leftrightarrow$ Python & 901 & 213 & 426 & 1540 & C $\leftrightarrow$ PHP & 296 & 51 & 102 & 449 \\
C $\leftrightarrow$ VB & 617 & 97 & 194 & 908 & C++ $\leftrightarrow$ C\# & 748 & 79 & 160 & 987 \\
C++ $\leftrightarrow$ Go & 792 & 208 & 418 & 1418 & C++ $\leftrightarrow$ Java & 753 & 172 & 345 & 1270 \\
C++ $\leftrightarrow$ Python & 842 & 202 & 405 & 1449 & C++ $\leftrightarrow$ PHP & 291 & 53 & 106 & 450 \\
C++ $\leftrightarrow$ VB & 586 & 97 & 195 & 888 & C\# $\leftrightarrow$ Go & 777 & 100 & 202 & 1079 \\
C\# $\leftrightarrow$ Java & 750 & 86 & 174 & 1010 & C\# $\leftrightarrow$ Python & 813 & 99 & 199 & 1111 \\
C\# $\leftrightarrow$ PHP & 293 & 40 & 80 & 413 & C\# $\leftrightarrow$ VB & 597 & 70 & 142 & 809 \\
Java $\leftrightarrow$ Go & 793 & 221 & 443 & 1457 & Java $\leftrightarrow$ Python & 838 & 217 & 436 & 1491 \\
Java $\leftrightarrow$ PHP & 574 & 119 & 239 & 932 & Java $\leftrightarrow$ VB & 610 & 104 & 210 & 924 \\
Go $\leftrightarrow$ Python & 887 & 314 & 628 & 1828 & Go $\leftrightarrow$ PHP & 606 & 128 & 258 & 992 \\
Go $\leftrightarrow$ VB & 618 & 116 & 232 & 966 & PHP $\leftrightarrow$ Python & 927 & 185 & 370 & 1482 \\
PHP $\leftrightarrow$ VB & 267 & 44 & 88 & 399 & VB $\leftrightarrow$ Python & 644 & 114 & 229 & 987 \\
    \bottomrule
  \end{tabular}
  \caption{Composition and Distribution of the Multilingual Dataset. The numbers refer to the number of code pair samples. VB is short for Visual Basic. [Return to Section \ref{sec:multilingualTrans_dataset}]}
  \label{table:Multilingual}
\end{table*}
}

\subsubsection{Code Translation Metrics}
\label{sec:code_translation_metrics}

\paragraph{Match-Based Evaluation Metrics} These evaluation metrics are based on the similarity between the translation output and the reference translation. Among them, the Exact Match (EM) metric calculates the percentage of translation outputs that \textit{exactly} match the reference translation, which overlooks the fact that the same function can be implemented in various ways. The Bilingual Evaluation Understudy (BLEU)~\cite{papineni2002bleu} metric evaluates the similarity between the translation output and the reference translation by multiplying the geometric average of n-gram precision scores with a brevity penalty. The CodeBLEU~\cite{ren2020codebleu} metric extends BLEU by considering syntactic and semantic characteristics of programming languages; it not only considers shallow matching but also pays attention to syntactic and semantic matching. CrystalBLEU~\cite{eghbali2022crystalbleu} focuses more on the inherent differences between source code and natural language, such as trivial shared n-gram syntax.   CodeBERTScore~\cite{zhou2023codebertscore} uses pre-trained models to encode the translation output and reference translation, then calculates the dot product similarity between them, enabling comparisons of code pairs with distinct lexical forms. However, CodeBLEU, CrystalBLEU, and CodeBERTScore have limitations as they only support a limited range of programming languages and cannot be used in general multilingual scenarios.  Ruby~\cite{tran2019does}, a new method for evaluating code translation, considers the lexical, syntactic, and semantic representations of source code. However, its codebase has not yet been open-sourced. These match-based evaluation metrics can only evaluate the surface form and semantic differences of the code, while neglecting the executability of the code and the functional equivalence of implementation variations.

\paragraph{Execution-Based Evaluation Metrics} Execution-based evaluation metrics mainly compare the executed result of the generated code with the expected result. The PASS@k score~\cite{kulal2019spoc} is evaluated by unit tests: if any of the $k$ samples meets the expected result, the generated result is deemed successful. AvgPassRatio~\cite{hao2022aixbench,hendrycks2021measuring} evaluates the overall executable result of code by calculating the average pass rate of test cases.  Computational accuracy~\cite{lachaux2020unsupervised} measures the quality of the generated code snippet by comparing the output of this snippet with the reference code snippet when given the same input. Additionally, CodeScore~\cite{dong2023codescore} claims that it can estimate the PassRatio of test cases for the generated code without executing the code, but its codebase has not yet been open-sourced. These execution-based evaluation metrics require construction of executable test sets, which could be costly. Furthermore, due to potential security threats from the execution environment and the code, they need to be run in an isolated sandbox.

\subsection{Data Management}
\label{sec:data_management}

\subsubsection{Data Sources \& Licenses}
\label{sec:data_sources_licenses}
We collect CodeTransOcean from two different platforms. The MultilingualTrans and NicheTrans datasets are collected from Rosetta Code\footnote{\url{https://rosettacode.org/wiki/Rosetta_Code}}, a programming site presenting solution strategies for identical tasks across as many programming languages as possible, thereby demonstrating both similarities and differences among these languages. We strictly adhere to the data distribution license of the platform as Attribution-ShareAlike 4.0 International (CC BY-SA 4.0) license\footnote{\url{https://creativecommons.org/licenses/by-sa/4.0/}}. 

The DLTrans dataset is derived from an open-source teaching platform \textit{Dive into Deep Learning}\footnote{\url{https://github.com/d2l-ai/d2l-zh}}, which is dedicated to teaching deep learning knowledge ranging from theoretical background, conceptual understanding, to coding practices. We strictly adhere to the data distribution license of this platform as Apache-2.0\footnote{\url{https://github.com/d2l-ai/d2l-zh/blob/master/LICENSE}}. To ensure legal and regulated use of these datasets, we require strict adherence to these licenses.

\subsubsection{Data Processing}
\label{sec:data_processing}

\paragraph{Multilingual Datasets} Given the variations in compilation requirements among programming languages, we keep the original format as much as possible to ensure the compilability of the data while ensuring its accuracy. Additionally, we employ a duplicate-file-detection tool to identify and remove duplicate data from the dataset to avoid any potential data leakage problems during model training.

\paragraph{Cross-Framework Dataset} To ensure the compilability of Python, we keep the original formatting information. We manually verify all automatically collected samples, identify and exclude samples that do not meet the requirements.

\subsubsection{Data Quality}
\label{sec:data_quality}
We randomly select 1K samples from each dataset within CodeTransOcean for manual quality assessment. We find that for the MultilingualTrans dataset with compilation requirements, the compilability rate exceeds \textbf{90\%}. We verify that the code pairs in each dataset are functionally identical and confirm that CodeTransOcean is of high quality.

Additionally, during data collection, we pay special attention to the diversity of domain knowledge and code styles. CodeTransOcean includes various code examples ranging from basic syntactic structures to complex algorithm implementations, as well as building neural networks from scratch and conducting training and inference. This rich diversity ensures that CodeTransOcean reflects a wide variety of real-world scenarios.

\begin{table*}[ht]
  \centering
  \renewcommand{\arraystretch}{1}
  \begin{tabular}{ccccc|ccccc}
    \toprule
    \textbf{Method} & \textbf{Train} & \textbf{Dev} & \textbf{Test} & \textbf{Total} & \textbf{Method} & \textbf{Train} & \textbf{Dev} & \textbf{Test} & \textbf{Total}\\
    \midrule
Ada         & 6022  & 464 & 937 & 7423  & Elixir      & 3618  & 297 & 599  & 4514  \\
Arturo      & 3802  & 470 & 947 & 5219  & Erlang      & 3269  & 217 & 449  & 3935  \\
AutoHotKey  & 4305  & 555 & 1120 & 5980  & Factor      & 4724  & 860 & 1756 & 7340  \\
AWK         & 3880  & 578 & 1162 & 5620  & Forth       & 3619  & 339 & 690  & 4648  \\
BBC Basic  & 3663  & 239 & 485  & 4387  & Fortran     & 4050  & 305 & 617  & 4972  \\
Clojure     & 3617  & 310 & 633  & 4560  & Groovy      & 3773  & 227 & 467  & 4467  \\
Common Lisp & 5085  & 467 & 933 & 6485  & Haskell     & 5647  & 1045 & 2097 & 8789  \\
D           & 5541  & 825 & 1662 & 8028  & Icon        & 2646  & 158 & 326  & 3130  \\
Delphi      & 3547  & 436 & 889  & 4872  & J           & 5476  & 1204 & 2422 & 9102  \\
Julia       & 5829  &1511 & 3055 & 10395 & Ruby        & 5793  & 1290 & 2600 & 9683  \\
Lua         & 5316  & 677 & 1366 & 7359  & COBOL       & 2438  & 167 & 355  & 2960  \\
Mathematica & 5485  & 1046 & 2105 & 8636  & REXX        & 5595  & 1118 & 2241 & 8954  \\
MATLAB      & 2872  & 157 & 322  & 3351  & R           & 2803  & 197 & 402  & 3402  \\
Nim         & 5814  & 1321 & 2675 & 9810  & Racket      & 5646  & 901 & 1817 & 8364  \\
OCaml       & 4286  & 405 & 817 & 5508  & Rust        & 5146  & 717 & 1439 & 7302  \\
Pascal      & 3393  & 465 & 942  & 4800  & Tcl         & 5354  & 740 & 1502 & 7596  \\
Perl        & 5818  &1445 & 2914 & 10177 & PowerShell  & 3563  & 240 & 490  & 4293  \\
Scala       & 5852  & 1074 & 2164 & 9090  & F\#          & 4517  & 638 & 1287 & 6442  \\
Swift       & 3653  & 404 & 818  & 4875 \\   

\bottomrule
  \end{tabular}
  \caption{The number of code samples for each language in the NicheTrans datasets. [Return to Section \ref{sec:nicheTrans_dataset}]}
  \label{table:NicheTrans}
\end{table*}

\subsection{Specific Challenges in Implementing Cross-framework Translation}
\label{appendix:specific_challenges}

Firstly, there are significant design differences between frameworks, including data processing methods, model-building strategies, and network connection techniques. Secondly, the inherent complexity of DL code increases the difficulty of conversion, as these codes usually contain various components such as neural network layers, loss functions, optimizers, and learning rate schedulers. Thirdly, there are significant inconsistencies in the code structure of different frameworks, such as code organization and variable naming rules. Lastly, cross-platform compatibility must be considered because DL code may encounter compatibility issues when executing on different hardware platforms (e.g., GPUs, CPUs, TPUs) and operating systems.

\subsection{Code Examples on Different Deep Learning Frameworks}
\label{appendix:dl_examples}
Figures \ref{fig:softmax_cross_entropy_loss} and \ref{fig:seq2seq_decoder} show the implementation of two different deep learning components in various deep learning frameworks.

\begin{figure}[ht]
  \centering
  \begin{minipage}{.48\textwidth}
  \textbf{PyTorch}
  \begin{lstlisting}[language=Python]
class MaskedSoftmaxCELoss(nn.CrossEntropyLoss):
    def forward(self, pred, label, valid_len):
        weights = torch.ones_like(label)
        weights = sequence_mask(weights, valid_len)
        self.reduction='none'
        unweighted_loss = super(MaskedSoftmaxCELoss, self).forward(
            pred.permute(0, 2, 1), label)
        weighted_loss = (unweighted_loss * weights).mean(dim=1)
        return weighted_loss
  \end{lstlisting}
  \end{minipage}\hspace{0.4cm}
  \begin{minipage}{.48\textwidth}
  \textbf{TensorFlow}
  \begin{lstlisting}[language=Python]
class MaskedSoftmaxCELoss(tf.keras.losses.Loss):
    def __init__(self, valid_len):
        super().__init__(reduction='none')
        self.valid_len = valid_len

    def call(self, label, pred):
        weights = tf.ones_like(label, dtype=tf.float32)
        weights = sequence_mask(weights, self.valid_len)
        label_one_hot = tf.one_hot(label, depth=pred.shape[-1])
        unweighted_loss = tf.keras.losses.CategoricalCrossentropy(
            from_logits=True, reduction='none')(label_one_hot, pred)
        weighted_loss = tf.reduce_mean((unweighted_loss*weights), axis=1)
        return weighted_loss
  \end{lstlisting}
  \end{minipage}
    \begin{minipage}{.48\textwidth}
  \textbf{MXNet}
  \begin{lstlisting}[language=Python]
class MaskedSoftmaxCELoss(gluon.loss.SoftmaxCELoss):
    def forward(self, pred, label, valid_len):
        weights = np.expand_dims(np.ones_like(label), axis=-1)
        weights = npx.sequence_mask(weights, valid_len, True, axis=1)
        return super(MaskedSoftmaxCELoss, self).forward(pred, label, weights)
  \end{lstlisting}
  \end{minipage}\hspace{0.4cm}
    \begin{minipage}{.48\textwidth}
  \textbf{Paddle}
  \begin{lstlisting}[language=Python]
class MaskedSoftmaxCELoss(nn.CrossEntropyLoss):
    def forward(self, pred, label, valid_len):
        weights = paddle.ones_like(label)
        weights = sequence_mask(weights, valid_len)
        self.reduction='none'
        unweighted_loss = super(MaskedSoftmaxCELoss, self).forward(
            pred, label)
        weighted_loss = (unweighted_loss * weights).mean(axis=1)
        return weighted_loss
  \end{lstlisting}
  \end{minipage}\hspace{0.4cm}
  \caption{\textit{Softmax cross-entropy loss function with masking} in PyTorch, TensorFlow, MXNet and Paddle.}
  \label{fig:softmax_cross_entropy_loss}
\end{figure}

\begin{figure*}[ht]
  \centering
  \begin{minipage}{.48\textwidth}
  \textbf{PyTorch}
  \begin{lstlisting}[language=Python]
class Seq2SeqDecoder(d2l.Decoder):
    def __init__(self, vocab_size, embed_size, num_hiddens, num_layers, dropout=0, **kwargs):
        super(Seq2SeqDecoder, self).__init__(**kwargs)
        self.embedding = nn.Embedding(vocab_size, embed_size)
        self.rnn = nn.GRU(embed_size + num_hiddens, num_hiddens, num_layers, dropout=dropout)
        self.dense = nn.Linear(num_hiddens, vocab_size)

    def init_state(self, enc_outputs, *args):
        return enc_outputs[1]

    def forward(self, X, state):
        X = self.embedding(X).permute(1, 0, 2)
        context = state[-1].repeat(X.shape[0], 1, 1)
        X_and_context = torch.cat((X, context), 2)
        output, state = self.rnn(X_and_context, state)
        output = self.dense(output).permute(1, 0, 2)
        return output, state
  \end{lstlisting}
  \end{minipage}\hspace{0.4cm}
  \begin{minipage}{.48\textwidth}
  \textbf{TensorFlow}
  \begin{lstlisting}[language=Python]
class Seq2SeqDecoder(d2l.Decoder):
    def __init__(self, vocab_size, embed_size, num_hiddens, num_layers, dropout=0, **kwargs):
        super().__init__(**kwargs)
        self.embedding = tf.keras.layers.Embedding(vocab_size, embed_size)
        self.rnn = tf.keras.layers.RNN(tf.keras.layers.StackedRNNCells(
            [tf.keras.layers.GRUCell(num_hiddens, dropout=dropout)
             for _ in range(num_layers)]), return_sequences=True, return_state=True)
        self.dense = tf.keras.layers.Dense(vocab_size)

    def init_state(self, enc_outputs, *args):
        return enc_outputs[1]

    def call(self, X, state, **kwargs):
        X = self.embedding(X)
        context = tf.repeat(tf.expand_dims(state[-1], axis=1), repeats=X.shape[1], axis=1)
        X_and_context = tf.concat((X, context), axis=2)
        rnn_output = self.rnn(X_and_context, state, **kwargs)
        output = self.dense(rnn_output[0])
        return output, rnn_output[1:]
  \end{lstlisting}
  \end{minipage}
    \begin{minipage}{.48\textwidth}
  \textbf{MXNet}
  \begin{lstlisting}[language=Python]
class Seq2SeqDecoder(d2l.Decoder):
    def __init__(self, vocab_size, embed_size, num_hiddens, num_layers, dropout=0, **kwargs):
        super(Seq2SeqDecoder, self).__init__(**kwargs)
        self.embedding = nn.Embedding(vocab_size, embed_size)
        self.rnn = rnn.GRU(num_hiddens, num_layers, dropout=dropout)
        self.dense = nn.Dense(vocab_size, flatten=False)

    def init_state(self, enc_outputs, *args):
        return enc_outputs[1]

    def forward(self, X, state):
        X = self.embedding(X).swapaxes(0, 1)
        context = state[0][-1]
        context = np.broadcast_to(context, (X.shape[0], context.shape[0], context.shape[1]))
        X_and_context = np.concatenate((X, context), 2)
        output, state = self.rnn(X_and_context, state)
        output = self.dense(output).swapaxes(0, 1)
        return output, state
  \end{lstlisting}
  \end{minipage}\hspace{0.4cm}
    \begin{minipage}{.48\textwidth}
  \textbf{Paddle}
  \begin{lstlisting}[language=Python]
class Seq2SeqDecoder(d2l.Decoder):
    def __init__(self, vocab_size, embed_size, num_hiddens, num_layers, dropout=0, **kwargs):
        super(Seq2SeqDecoder, self).__init__(**kwargs)
        self.embedding = nn.Embedding(vocab_size, embed_size)
        weight_attr = paddle.ParamAttr(initializer=nn.initializer.XavierUniform())
        weight_ih_attr = paddle.ParamAttr(initializer=nn.initializer.XavierUniform())
        weight_hh_attr = paddle.ParamAttr(initializer=nn.initializer.XavierUniform())
        self.rnn = nn.GRU(embed_size + num_hiddens, num_hiddens, num_layers, dropout=dropout,
                          time_major=True, weight_ih_attr=weight_ih_attr,weight_hh_attr=weight_hh_attr)
        self.dense = nn.Linear(num_hiddens, vocab_size,weight_attr=weight_attr)

    def init_state(self, enc_outputs, *args):
        return enc_outputs[1]

    def forward(self, X, state):
        X = self.embedding(X).transpose([1, 0, 2])
        context = state[-1].tile([X.shape[0], 1, 1])
        X_and_context = paddle.concat((X, context), 2)
        output, state = self.rnn(X_and_context, state)
        output = self.dense(output).transpose([1, 0, 2])
        return output, state
  \end{lstlisting}
  \end{minipage}\hspace{0.4cm}
  \caption{Implementing \textit{RNN Decoder for Seq2Seq Learning} in PyTorch, TensorFlow, MXNet and Paddle. [Return to Section \ref{sec:DLTask}]}
  \label{fig:seq2seq_decoder}
\end{figure*}

\subsection{Multilingual Modeling}
\label{appendix:multilingual_modeling}

\paragraph{One-to-One}~For each language pair in the dataset, we train an independent model, e.g., translating C++ to Java.

\paragraph{One-to-Many} We train individual models from one language to many other languages, e.g., translating Python to all other languages.

\paragraph{Many-to-One} We train individual models from multiple languages to one language, e.g., translating all other languages to Python.

\paragraph{Many-to-Many} We train a unified model for the multiple to multiple languages in the dataset, which can handle translations between all languages.

We ensure all experiments are performed under the same hyperparameters and environment for comparison. Table \ref{tab:codet5+_parameters} shows these in detail.

\begin{table}[ht]
\centering
\begin{adjustbox}{width=0.48\textwidth}
\begin{tabular}{cccc}
\toprule
  & \textbf{MultilingualTrans} & \textbf{NicheTrans} & \textbf{DLTrans} \\ 
\midrule
Learning rate & 3e-5 & 2e-5 & 3e-5 \\
Beam size & 1 & 1 & 5  \\
Max source length & 1536 & 1536 & 512 \\
Max target length & 1536 & 1536 &  512 \\
Batch size & \multicolumn{3}{c}{16}   \\
Max epoch & \multicolumn{3}{c}{5}   \\
Fp16 & \multicolumn{3}{c}{True}   \\
GPU & \multicolumn{3}{c}{NVIDIA Tesla V100 32GB} \\
\bottomrule
\end{tabular}
\end{adjustbox}
\caption{Parameters and hardware configuration for training CodeT5+\_220M (220M is the model size) on CodeTransOcean.}
\label{tab:codet5+_parameters}
\end{table}

\begin{table*}[ht]
\centering
\begin{adjustbox}{width=\textwidth}
\begin{tabular}{ccccccccccc}
\toprule
& Method  & C & C++ & C\# & Go & VB & Python & Java & PHP \\ \midrule
\multirow{5}{*}{C}  
    &  Naive  & --  & 14.61 & 9.08 & 4.52 & 1.64 & 2.17 & 11.23 & 2.09  \\
    &  OtO  & --  & 11.13$\pm$0.28 & 4.77$\pm$0.85 & 8.18$\pm$0.40 & 2.02$\pm$0.23 & 1.85$\pm$0.14 & 6.50$\pm$2.91  & 1.71$\pm$0.32  \\
    &  OtM  & --  & 12.33$\pm$0.58 & 7.56$\pm$0.98 & 9.58$\pm$0.74 & 2.22$\pm$0.89 & 2.82$\pm$0.58 & 8.92$\pm$0.67  & 3.42$\pm$0.09  \\
    &  MtO  & --  & 7.99$\pm$1.94 & 5.35$\pm$0.41 & 6.94$\pm$0.83 & 3.77$\pm$0.44 & 2.09$\pm$0.25 & 8.65$\pm$0.52  & 1.93$\pm$0.49  \\ 
    &  MtM  & --  & 9.61$\pm$0.32 & 7.15$\pm$1.41 & 7.54$\pm$0.33 & 2.30$\pm$0.71 & 2.11$\pm$0.43 & 8.29$\pm$0.45  & 3.30$\pm$0.92  \\ 
\hdashline
\multirow{5}{*}{C++}    
    &  Naive  & 14.88  & -- & 10.08 & 3.87 & 3.76 & 1.69 & 11.39  & 1.92  \\
    &  OtO   & 10.58$\pm$0.37  & -- & 6.52$\pm$1.96 & 6.77$\pm$1.08 & 2.34$\pm$0.90  & 1.82$\pm$0.37 & 9.41$\pm$1.59  & 1.50$\pm$0.03  \\ 
    &  OtM   & 13.32$\pm$2.46  & -- & 9.87$\pm$1.15 & 9.83$\pm$1.37 & 2.62$\pm$1.15  & 2.93$\pm$0.18 & 12.10$\pm$1.51  & 2.87$\pm$0.82  \\ 
    &  MtO   & 10.54$\pm$2.33  & -- & 6.98$\pm$1.46 & 7.29$\pm$0.73 & 3.57$\pm$0.74  & 2.28$\pm$0.18 & 7.82$\pm$1.26  & 1.96$\pm$0.20  \\
    &  MtM   & 9.92$\pm$1.14  & -- & 7.79$\pm$1.49 & 7.70$\pm$0.48 & 2.06$\pm$0.74  & 2.38$\pm$0.30 & 10.20$\pm$1.20 & 4.12$\pm$0.81  \\
\hdashline
\multirow{5}{*}{C\#}
  &  Naive  & 9.05 & 10.03 & -- & 5.05 & 7.20 & 1.83 & 13.60 & 2.58  \\
  &   OtO  & 5.41$\pm$1.25  & 6.45$\pm$1.68 &  -- & 7.10$\pm$0.63 & 9.42$\pm$5.50 & 1.73$\pm$0.46 & 10.55$\pm$1.67 & 1.97$\pm$0.76  \\ 
  &   OtM  & 7.42$\pm$1.82  & 8.99$\pm$0.17 &  -- & 9.81$\pm$0.68 & 6.13$\pm$2.92 & 2.80$\pm$0.43 & 11.92$\pm$2.52 & 4.55$\pm$0.81  \\
  &   MtO  & 6.37$\pm$0.69  & 6.13$\pm$1.60 & --  & 7.53$\pm$0.94 & 9.92$\pm$1.00 & 2.47$\pm$0.62 & 10.16$\pm$1.40 & 1.70$\pm$0.20  \\  
  &   MtM  & 6.19$\pm$0.88  & 7.22$\pm$0.93 & --  & 9.36$\pm$0.92 & 4.60$\pm$2.87 & 2.23$\pm$0.36 & 12.50$\pm$1.35 & 4.70$\pm$0.27 \\ 
\hdashline
\multirow{5}{*}{Go}    
    &  Naive  & 4.52 & 3.75 & 5.04 & -- & 2.46 & 3.00 & 6.56 & 2.14  \\
    &  OtO   & 5.65$\pm$0.41  & 6.88$\pm$0.47 & 4.70$\pm$1.28 & -- & 1.87$\pm$0.54 & 2.89$\pm$0.16  & 5.41$\pm$1.06  & 2.48$\pm$0.23  \\ 
    &  OtM   & 6.06$\pm$0.99  & 6.56$\pm$0.89 & 7.28$\pm$0.82 & -- & 2.12$\pm$0.65 & 3.32$\pm$0.36  & 8.79$\pm$1.06  & 3.48$\pm$0.20  \\ 
    &  MtO   & 5.41$\pm$0.63  & 4.77$\pm$0.76 & 6.89$\pm$0.33 & -- & 2.94$\pm$0.63 & 3.12$\pm$0.48  & 7.81$\pm$0.32  & 1.79$\pm$0.51  \\
    &  MtM   & 5.18$\pm$1.23  & 6.06$\pm$0.28 & 7.12$\pm$0.72 & -- & 1.82$\pm$0.93 & 2.83$\pm$0.50  & 8.99$\pm$1.28  & 2.55$\pm$0.40  \\
\hdashline
\multirow{5}{*}{VB}    
    &  Naive  & 1.64 & 3.76 & 7.29 & 2.54 & -- & 1.42 & 2.89  & 0.46  \\
    & OtO   & 3.96$\pm$0.20  & 5.25$\pm$0.68 & 17.34$\pm$1.79 & 5.85$\pm$0.64 & -- & 1.34$\pm$0.06 & 5.51$\pm$0.43 & 0.61$\pm$0.18 \\
    &  OtM  & 5.15$\pm$0.15  & 6.10$\pm$0.30 & 19.63$\pm$1.03 & 7.83$\pm$0.45 & -- & 2.02$\pm$0.32 & 7.91$\pm$0.92 & 2.02$\pm$0.20 \\
    &  MtO  & 4.25$\pm$0.86  & 4.15$\pm$1.09 & 11.96$\pm$1.96 & 6.38$\pm$0.45 & -- & 1.66$\pm$0.58 & 7.26$\pm$1.39 & 1.18$\pm$0.33 \\
    &  MtM  & 5.18$\pm$0.65  & 5.09$\pm$0.15 & 14.13$\pm$1.83 & 6.97$\pm$1.30 & -- & 2.34$\pm$0.11 & 8.39$\pm$0.77 & 2.92$\pm$0.22 \\
\hdashline
\multirow{5}{*}{Python} 
    &  Naive  & 1.73  & 1.14 & 1.44 & 2.50 & 0.89 & -- &  1.82 & 1.28  \\
    &  OtO  & 4.35$\pm$0.44 & 3.66$\pm$0.95 & 5.38$\pm$0.76 & 5.67$\pm$0.34 & 2.53$\pm$1.11 & -- & 4.70$\pm$0.07 & 3.24$\pm$0.60 \\ 
    &  OtM  & 4.51$\pm$1.33  & 5.18$\pm$0.52 & 6.48$\pm$0.71 & 6.04$\pm$0.96 & 1.50$\pm$0.56 & -- & 6.29$\pm$0.39 & 6.55$\pm$0.94 \\ 
    &  MtO  & 4.70$\pm$0.28  & 3.44$\pm$0.76 & 5.48$\pm$0.55 & 4.80$\pm$0.63 & 2.42$\pm$0.18 & -- & 5.32$\pm$0.66 & 2.74$\pm$0.39 \\
    &  MtM   & 4.54$\pm$0.67  & 4.74$\pm$0.14 & 5.69$\pm$0.72 & 5.84$\pm$0.36 & 2.09$\pm$0.94 & -- & 6.45$\pm$0.11 & 4.49$\pm$0.63 \\
\hdashline
\multirow{5}{*}{Java}  
    &  Naive  & 11.23 & 11.15 & 13.44 & 6.57 & 2.84 & 2.24 & --  & 2.60  \\
    & OtO & 7.15$\pm$1.59 & 9.09$\pm$1.00 & 10.92$\pm$1.61 & 11.05$\pm$0.52 & 2.66$\pm$0.13 & 2.54$\pm$0.35 & -- & 2.05$\pm$0.95 \\
    &  OtM & 9.27$\pm$1.52  & 9.33$\pm$0.86 & 13.52$\pm$0.95 & 12.57$\pm$0.96 & 2.92$\pm$0.78 & 3.98$\pm$0.86 & -- & 6.20$\pm$1.08 \\ 
    &  MtO  & 6.69$\pm$0.83  & 6.32$\pm$1.66 & 10.57$\pm$1.16 & 8.79$\pm$0.44 & 3.91$\pm$0.57 & 2.86$\pm$0.36 & -- & 2.48$\pm$0.34 \\
    &  MtM  & 6.56$\pm$0.33  & 7.60$\pm$0.46 & 9.40$\pm$1.08 & 8.30$\pm$0.87 & 2.20$\pm$0.85 & 2.53$\pm$0.15 & -- & 3.57$\pm$0.94 \\
\hdashline
\multirow{5}{*}{PHP}  
    &  Naive  & 1.79 & 1.51 & 2.36 & 2.00 & 0.37 & 1.30 & 2.30  & --  \\
    & OtO    & 4.18$\pm$1.44  & 2.91$\pm$0.43 & 6.45$\pm$0.43 & 5.85$\pm$0.54 & 0.42$\pm$0.09 & 1.98$\pm$0.26 & 6.47$\pm$0.59 & -- \\ 
    &  OtM   & 3.78$\pm$1.80  & 1.93$\pm$1.35 & 5.51$\pm$0.60 & 5.13$\pm$0.60 & 0.56$\pm$0.36 & 2.70$\pm$0.80 & 5.51$\pm$1.91 & -- \\ 
    &  MtO   & 4.97$\pm$1.09  & 3.69$\pm$0.46 & 6.08$\pm$3.03 & 6.44$\pm$0.78 & 2.28$\pm$0.12 & 3.55$\pm$0.18 & 8.66$\pm$0.84 & -- \\
    &  MtM   & 6.03$\pm$0.61  & 5.15$\pm$0.67 & 10.64$\pm$1.27 & 6.69$\pm$0.31 & 1.69$\pm$0.57 & 2.46$\pm$0.46 & 7.87$\pm$1.90 & -- \\
\bottomrule
\end{tabular}
\end{adjustbox}
\caption{BLEU scores from different multilingual modeling strategies by fine-tuning the pre-trained CodeT5+\_220M model (220M is the model size)~\cite{wang2023codet5}. Naive denotes Naive Copy, which directly duplicates the source code as the target code without making any modifications. Method OtO, OtM, MtO, and MtM denote
One-to-One, One-to-Many, Many-to-One, and Many-to-Many, respectively. 
The rows correspond to the source language while the columns correspond to the target language.
We run each experiment with three different random seeds and report the mean and standard deviation of BLEU scores. [Return to Section~\ref{sec:multilingual_modeling}.]}
\label{tab:multilingual_modeling_strategies}
\end{table*}

\subsection{Prompt Variations}
\label{appendix:zero-shot}

\paragraph{Role Assignment}\cite{peng2023making,alkhamissi2023optr,wu2023large,f2023awesome} We configured two distinct roles for the model, each with unique skills. This arrangement empowers the model to simulate more domain-adaptable and specialized expert roles.

\paragraph{Polite inquiry}\cite{f2023awesome} These strategies add polite expression and set up imperative and interrogative requests. Given that ChatGPT is designed to simulate human conversation styles as closely as possible, including understanding and simulating polite language expressions. Therefore, we expect these strategies to boost the comprehension of the model and augment the quality of its generated results.
 
\paragraph{Clarify usage} This strategy aims to make the model clearly aware of its requirements during the code translation process - the generated code needs to be guaranteed to execute without issues.

\noindent The translation prompts of the above four strategies are shown in Table \ref{tab:zeroshot_prompts}.

\begin{table*}[ht]
\centering
\setlength{\tabcolsep}{4pt}
\resizebox{1.5\columnwidth}{!}{
\begin{tabular}{l p{8cm}}
\toprule
\bf Method & \multicolumn{1}{c}{\bf Translation Prompt} \\
\midrule
\textbf{Role Assignment \#1} & \texttt{"role": "system", "content": "Your are a code translation system.", "role": "user", "content": "Please provide the \useblue{[TL]} translation for the following \useblue{[SL]} code:\useblue{[SC]} }\\
\hdashline
\textbf{Role Assignment \#2} & \texttt{"role": "system", "content": "You are a code translation system that specializes in \useblue{[SL]} and \useblue{[TL]} programming languages.", "role": "user", "content": "Please provide the \useblue{[TL]} translation for the following \useblue{[SL]} code:\useblue{[SC]} }\\
\hdashline
\textbf{Role Assignment \#3} & \texttt{"role": "system", "content": "You are a programmer proficient in multiple programming languages.", "role": "user", "content": "Please provide the \useblue{[TL]} translation for the following \useblue{[SL]} code:\useblue{[SC]} }\\
\hdashline
\textbf{Role Assignment \#4} & \texttt{"role": "system", "content": "You are a programmer proficient in \useblue{[SL]} and \useblue{[TL]} programming languages.", "role": "user", "content": "Please provide the \useblue{[TL]} translation for the following \useblue{[SL]} code:\useblue{[SC]} }\\
\midrule
\textbf{Polite Inquiry \#1} & \texttt{Please translate the following \useblue{[SL]} code into \useblue{[TL]} code:\useblue{[SC]}} \\
\hdashline
\textbf{Polite Inquiry \#2} & \texttt{Can you rewrite this \useblue{[SL]} code in \useblue{[TL]}?\useblue{[SC]}} \\
\midrule
\textbf{Clarify Usage} & \texttt{Translating \useblue{[SL]} to \useblue{[TL]} ensures that Python code can be executed.\useblue{[SC]}} \\
\midrule
\textbf{Divide \& Conquer} & \texttt{Translate \useblue{[SL]} to \useblue{[TL]}:\useblue{[SC]}} \\
\bottomrule
\end{tabular}
}
  \caption{Translation prompts for prompt variants and contextual strategies on ChatGPT. [SL] refers to the source language, [SC] refers to the source code, [TL] refers to the target language. [Return to Section \ref{sec:chatgpt}]}
  \label{tab:zeroshot_prompts}
\end{table*}

\begin{table*}[ht]
    \begin{adjustbox}{width=\textwidth}
    \begin{tabular}{cccccc|cccccc}
    \toprule
        \textbf{Strategy} & \textbf{Temp.} & \textbf{EM} & \textbf{BLEU} & \textbf{CodeBLEU} & \textbf{DSR@0} & \textbf{Strategy} & \textbf{Top-K} & \textbf{EM} & \textbf{BLEU} & \textbf{CodeBLEU} & \textbf{DSR@0} \\
    \midrule
 \multirow{6}{*}{Top-K = 0}& 0 & 0.57 & \textbf{10.83} & 24.45 & 48.57\% & \multirow{6}{*}{Temp. = 0} & 0 & 0.57 & \textbf{10.83} & \textbf{24.45} & 48.57\%  \\
  & 0.2 & 0.57 & 10.82 & 24.43 & 47.71\% &  & 0.2 & 0.57 & 10.81 & \textbf{24.45} & 48.29\%  \\
 & 0.4 & 0.57 & 10.80 & 24.37 & 48.00\% &  & 0.4 & 0.57 & 10.81 & 24.44 & 48.29\%  \\
 & 0.6 & 0.57 & 10.78 & 24.45 & 48.00\% &  & 0.6 & 0.57 & 10.82 & 24.42 & 48.29\%  \\
  & 0.8 & 0.57 & 10.82 & \textbf{24.49} & 47.71\% &  & 0.8 & 0.57 & 10.80 & 24.38 & 47.71\% \\
 & 1.0 & 0.57 & 10.81 & 24.41 & 48.00\% & & 1.0 & 0.57 & 10.81 & 24.38 & 48.00\%  \\
    \bottomrule
  \end{tabular}
  \end{adjustbox}
    \caption{Code translation performance of ChatGPT under different parameter settings. \textit{Temp.} refers to temperature. [Return to Section \ref{sec:chatgpt}]}
  \label{table:chatgpt_parameters}
\end{table*}

\begin{table*}[ht]
    \centering
    \begin{tabular}{p{0.94\linewidth}}
    \toprule
\textbf{~Self-debug@0} \\
\vspace{-1mm}
\begin{flushright}
\linespread{0.8}\selectfont
Translate [source\_language] to [target\_language]: [source\_code]. ~~\user
\end{flushright}
\openai ~~Here is the [target\_language] code equivalent \\
~~~~~~~~of the given [source\_language] code: [translated\_code].  \\
\hdashline
\vspace{-1mm}
\textbf{~Self-debug@n}
\begin{flushright}
\linespread{0.8}\selectfont
The above python code executes with the following errors, ~~\user \\ please correct them. \color{red}{[Compiler reports errors]}~~~~~~~~~\\
\end{flushright}
\openai ~~Here is the modified [target\_language] code: [translated\_code].
\\ \bottomrule
\end{tabular}
\caption{A simple demo: Translation prompting of ChatGPT in the multi-round debugging strategy.\textcolor{red}{~The content in red is returned by the compiler.}  [Return to section \ref{sec:chatgpt}.]}
\label{tab:debug@k_prompts}
\end{table*}

\begin{table*}[ht]
    \centering
    \begin{tabular}{p{0.94\linewidth}}
    \toprule
\textbf{~Zero-shot prompting} \\
\vspace{-1mm}
\begin{flushright}
\linespread{0.8}\selectfont
Does the following Python code execute? [python\_code]. ~~\user
\end{flushright}
\openai ~~Yes, the Python code executes without errors. \\
\begin{flushright}
\linespread{0.8}\selectfont
Please predict the executed output of the Python code above. ~~\user \\ 
\end{flushright}
\openai ~~The predicted execution result of the Python code above is [output]. \\
 \midrule
\textbf{~Few-shot prompting} \\
\vspace{-1mm}
\begin{flushright}
\linespread{0.8}\selectfont
This is a executable Python code [python\_code], and this is a Python code [python\_code]~~\user \\ 
 that cannot be executed. Does the following Python code execute? [python\_code].~~~~~~~~~\\
\end{flushright}
\openai ~~Yes, the Python code executes without errors. \\
\begin{flushright}
\linespread{0.8}\selectfont
Please predict the executed output of the Python code above. ~~\user \\ 
\end{flushright}
\openai ~~The predicted execution result of the Python code above is [output].\\
\bottomrule
\end{tabular}
\caption{Two simple demos: prompting in fuzzy execution experiments. [Return to Section \ref{sec:limitations}.]}
\label{tab:fuzzy_prompts}
\end{table*}

\subsection{One-Shot}
\label{appendix:one_shot}
\paragraph{Experiment \#1} selects a training sample from a high-resource language code pair as an example. In this case, the target language type aligns with the target language type of the translation request, but the source language does not.
 
\paragraph{Experiment \#2} selects a code pair whose source and target language directions are congruent with the translation requirements as an example. That is, the source and target languages of the example dynamically adjust following the translation requirements.
 
\paragraph{Experiment \#3} randomly selects a code pair as an example, in which neither the source nor the target languages match the translation requirements.

\noindent The specific translation prompts are shown in Table \ref{table:oneshot_cot_prompts}.

\begin{table*}[ht]
\setlength{\tabcolsep}{4pt}
\centering
\resizebox{1.4\columnwidth}{!}{
\begin{tabular}{l p{8cm}}
\toprule
\bf Method & \multicolumn{1}{c}{\bf Translation Prompt} \\
\midrule
\textbf{One-Shot} & \texttt{Here is an example of a translation from [ESL] to [ETL].[ESL]: [ESC], [ETL]: [ETC]. Please imitate this example to translate following code from [SL] to [TL]:[TC].} \\
\hdashline
\textbf{One-Shot \#1} & \texttt{where [ESL]$\neq$[SL] and [ETL]=[TL]} \\
\hdashline
\textbf{One-Shot \#2} & \texttt{where [ESL]=[SL] and [ETL]=[TL]}\\
\hdashline
\textbf{One-Shot \#3} & \texttt{where [ESL]$\neq$[SL] and [ETL]$\neq$[TL]}\\
\midrule
\textbf{CoT \#1} & \texttt{1.Please explain the function of the following [SL] code, which is limited to 200 words.[SC]
2.Please translate into [TL] code according to the following [SL] code and its functional description.[SL]:[SC].Function description:[DSC]}\\
\hdashline
\textbf{CoT \#2} & \texttt{First, understand the function of the following [SL] code. Then, translate the [SL] code into [TL] code while keeping the function unchanged.[SC]}\\
\hdashline
\textbf{CoT \#3} & \texttt{First, understand the functionality of the following [SL] code and predict the execution output. Then, translate the [SL] code into [TL] while maintaining the same functionality, ensuring that the translated code can be successfully executed.[SL]} \\
\hdashline
\textbf{CoT \#4} & \texttt{First, learn how to translate [ESL] code to [ETL] based on the example, [SL]:[ESC],[TL]:[ETC]. Then, understand the functionality of the following [SL] code and predict the execution output, [SL]:[SC]. Finally, translate the [SL] code into [TL] while maintaining the same functionality, ensuring that the translated code can be successfully executed.} \\
\bottomrule
\end{tabular}
}
  \caption{ChatGPT translation prompts on one-shot \& CoT strategies. [SL] refers to the source language, [SC] refers to the source code, [TL] refers to the target language,[ESL] refers to the source language in the example, [ESC] refers to the source code in the example, [ETL] refers to the target language in the example, [ETC] refers to the target code in the example, [DSC] refers to the natural language description of the source code. [Return to Section \ref{sec:chatgpt}]}
\label{table:oneshot_cot_prompts}

\end{table*}

\subsection{Chain of Thought}
\label{appendix:cot}

\paragraph{Experiment \#1} First, describe the function of the source code in the natural language, then translate it according to the source code and the corresponding natural language description.
 
\paragraph{Experiment \#2} First, let ChatGPT understand the function of the source code, followed by the translation, while ensuring that the function of the code remains unchanged during the translation process.
 
\paragraph{Experiment \#3} First, let ChatGPT understand the function of the source code, then predict the output result of the source code, and finally perform the translation, demanding that the translated code successfully executes.

\paragraph{Experiment \#4} ~Building upon Experiment \#3 and the one-shot approach of Experiment \#2, we introduce a CoT one-shot variation. That is, first, provide a case in the same direction for ChatGPT reference, then require it to understand the function of the source code, then predict the output of the source code, and finally translate it, with the condition that the translated code must successfully execute.

\noindent The specific translation prompts are shown in Table \ref{table:oneshot_cot_prompts}.

\end{document}